\documentclass[10pt,journal,compsoc]{IEEEtran}
\usepackage{epsfig}
\usepackage{graphicx}
\usepackage{amsmath}
\usepackage{amssymb}
\usepackage{bm}
\usepackage{algorithm}
\usepackage{algorithmic}
\usepackage{color}

\ifCLASSOPTIONcompsoc
  \usepackage[nocompress]{cite}
\else
  \usepackage{cite}
\fi
\ifCLASSINFOpdf
\else
\fi
\begin{document}
%
\title{Deep Differentiable Random Forests for Age Estimation}
%
%
%
%

\author{Wei~Shen, Yilu Guo, Yan Wang, Kai Zhao, Bo Wang,
        and Alan~Yuille,~\IEEEmembership{Fellow,~IEEE,}
\IEEEcompsocitemizethanks{\IEEEcompsocthanksitem W. Shen, Y. Wang and A. Yuille are with Department of Computer Science, Johns Hopkins University, Baltimore, MD 21218-2608 USA. \protect\\
E-mail: \{shenwei1231,wyanny.9,alan.l.yuille\}@gmail.com.
\IEEEcompsocthanksitem Y. Guo is with School of Communication and Information Engineering, Shanghai University, Shanghai 200444
China.
E-mail: gyl.luan0@gmail.com.
\IEEEcompsocthanksitem K. Zhao is with College of Computer and Control Engineering, Nankai University, Tianjin 300071, China.  \protect\\
E-mail: zhaok1206@gmail.com
\IEEEcompsocthanksitem B.Wang is with Vector Institute and Peter Munk Cardiac Center of University Health Network, Toronto, ON, Canada.  \protect\\
E-mail: bowang@vectorinstitute.ai}
\thanks{Manuscript received XX, 2018; revised XX, 2019.}}

%
%

\markboth{Journal of \LaTeX\ Class Files,~Vol.~XX, No.~XX, XX~2019}%
{Shen \MakeLowercase{\textit{et al.}}: Bare Demo of IEEEtran.cls for Computer Society Journals}
%



\IEEEtitleabstractindextext{%
\begin{abstract}
Age estimation from facial images is typically cast as a label distribution learning or regression problem, since aging is a
gradual progress. Its main challenge is the facial feature space w.r.t. ages is inhomogeneous, due to the large variation in facial
appearance across different persons of the same age and the non-stationary property of aging. In this paper, we propose two Deep Differentiable
Random Forests methods, Deep Label Distribution Learning Forest (DLDLF) and Deep Regression Forest (DRF), for age estimation.
Both of them connect split nodes to the top layer of convolutional neural networks (CNNs) and deal with inhomogeneous data by jointly
learning input-dependent data partitions at the split nodes and age distributions at the leaf nodes. This joint learning follows an
alternating strategy: (1) Fixing the leaf nodes and optimizing the split nodes and the CNN parameters by Back-propagation; (2) Fixing
the split nodes and optimizing the leaf nodes by Variational Bounding. Two Deterministic Annealing processes are introduced into the
learning of the split and leaf nodes, respectively, to avoid poor local optima and obtain better estimates of tree parameters free of initial
values. Experimental results show that DLDLF and DRF achieve state-of-the-art performance on three age estimation datasets.
\end{abstract}

}

\maketitle

\IEEEdisplaynontitleabstractindextext

%
\IEEEpeerreviewmaketitle

\IEEEraisesectionheading{\section{Introduction}\label{sec:intro}}

%
%
%
%
\IEEEPARstart{T}here has been a growing interest in age estimation from facial images, driven by the increasing demands for a variety of potential applications in forensic research~\cite{Ref:Alkass10}, security control~\cite{Ref:HanOJ13}, human-computer interaction (HCI)~\cite{Ref:HanOJ13} and social media~\cite{Ref:Rothe16}. In this paper, we focus on estimating
the precise chronological age (\emph{i.e.}, not age group estimation~\cite{Ref:LeviH15}). Although considerable progress has been made recently~\cite{Ref:ChenZDLR17,Ref:Agustsson17,Ref:Pan18}, estimating ages accurately and reliably from facial images is still a challenging problem.

To address age estimation, the characteristics of this task should be considered. First, aging is a slow and gradual
progress, thus there is a strong correlation between close ages of the same individual, \emph{e.g.}, a person's facial images taken at close ages are similar. Due to this fact, age estimation is usually formulated as a label distribution learning (LDL)~\cite{Ref:Geng2010Facial,Ref:Geng13,Ref:He17} or regression~\cite{Ref:GuoM11,Ref:Niu16,Ref:Huang17} problem rather than a classification problem, because in a classification problem, class labels are uncorrelated. Unlike classification, LDL assigns a distribution over the set of
labels to an instance, which can be obtained by fitting a Gaussian or Triangle distribution whose peak is the label of this instance and represents the relative importance of each
label involved in the description of an instance; by contrast, regression considers labels as continuous numerical values. Therefore, LDL and regression can explicitly and implicitly
model cross age correlations of the same individual, respectively.

\begin{figure}[!t]
\centering
\includegraphics[trim=0cm 6cm 0cm 7cm, clip=true, width=1.0\linewidth]{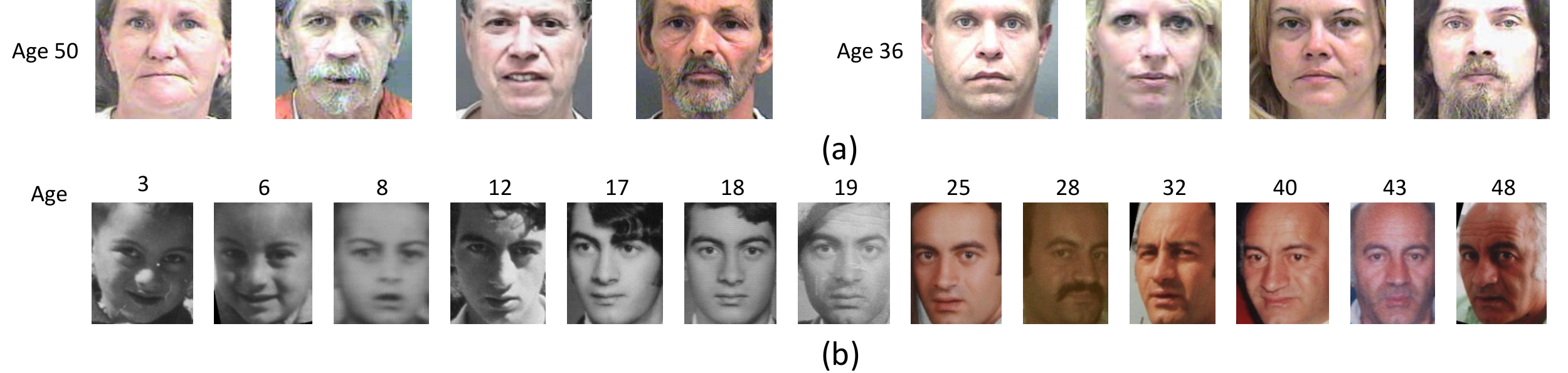}
\caption{(a) The large variation in facial appearance across different persons of the same age. (b) Facial images of a person from childhood to adulthood. Note that, Facial aging effects appear as changes in the shape of the face during childhood and changes in skin texture during adulthood, respectively.}\label{fig:Example}
\end{figure}

Second, learning the mapping between facial image features
and ages is challenging. The main difficulty is the facial
feature space w.r.t. ages is inhomogeneous, due to two factors:
1) there is a large variation in facial appearance across
different persons of the same age (Fig.~\ref{fig:Example}(a)); 2) the human face matures in different ways at different ages, \emph{e.g.}, bone growth in childhood and skin wrinkles in adulthood~\cite{Ref:RamanathanCB09} (Fig.~\ref{fig:Example}(b)). This inhomogeneity suggests applying divide-and-conquer models, such as Random Forests~\cite{Ref:Amit97,Ref:Breiman01,Ref:Shotton13}, to partition the data space and learn multiple local age estimators~\cite{Ref:HanOLJ15}. However, traditional Random Forests make hard data partitions based on heuristics, such as using a greedy algorithm where locally-optimal hard decisions are made at each split node~\cite{Ref:Amit97}, thus have limitations in representation learning, \emph{e.g.}, they can not learn deep facial features to perform data partition in an end-to-end manner.

To address this issue, we propose two Deep Differentiable Random
Forests for age estimation, where one is an LDL model,
named by Deep Label Distribution Learning Forest
(DLDLF), the other is a regression model, named by Deep
Regression Forest (DRF). Our Deep Differentiable Random Forests are
inspired by~\cite{Ref:Kontschieder15}, which introduced differentiable decision
classification trees and integrated them with CNNs by connecting
the split nodes in trees to a fully connected layer
of a CNN. We extend the differentiable trees to deal with
LDL and regression problems, which is non-trivial (see the
discussion in Sec.~\ref{sec:related_work}). Differentiable trees perform soft data
partition at split nodes, so that an input-dependent partition
function can be learned to handle inhomogeneous data.
In addition, the deep facial features at split nodes (input
feature space) and the age distributions at leaf nodes (local
estimators) can be learned jointly, which ensures that the
local input-output correlation is homogeneous at the leaf
node.

\begin{figure*}[!t]
\centering
\includegraphics[trim=0.5cm 0.5cm 0.5cm 0.5cm, clip=true, width=1.0\linewidth]{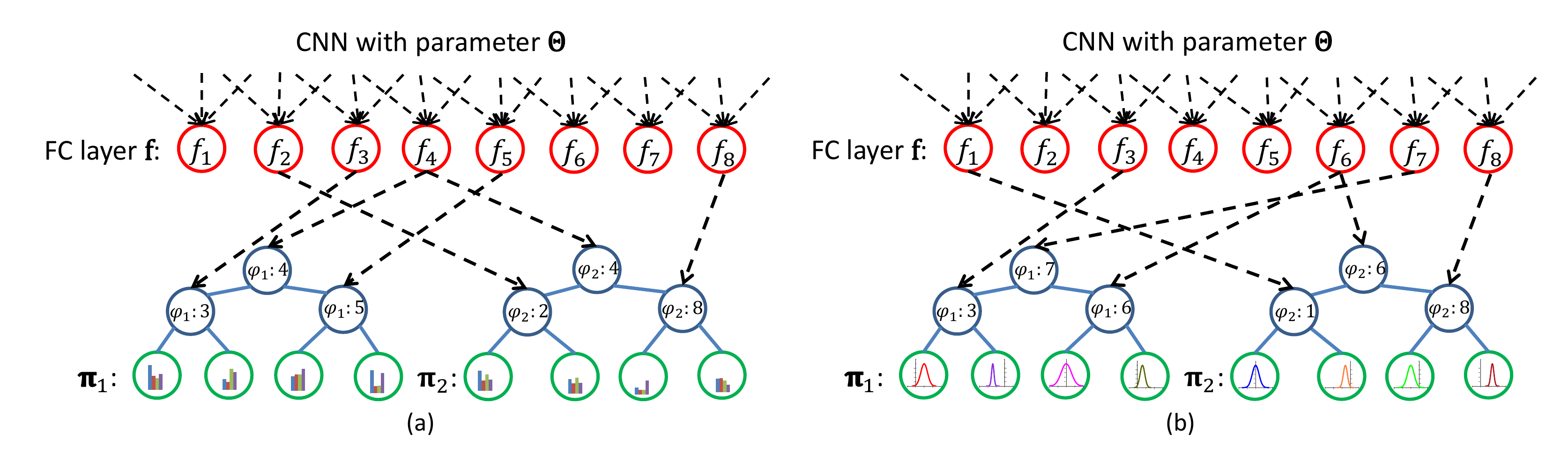}
\caption{Illustration of (a) a deep label distribution learning forest (DLDLF) and (b) a deep regression forest (DRF). Each forest consists of two trees. The top red circles denote the output units of the function $\mathbf{f}$ parameterized by $\bm{\Theta}$. Here, they are the units of a fully-connected (FC) layer in a CNN. The blue and green circles are split nodes and leaf nodes, respectively. in each forest, two index functions $\varphi_1$ and $\varphi_2$ are randomly assigned to the two trees respectively before training and then fixed. The black dash arrows indicate the correspondence between the split nodes of the two trees and the output units of the FC layer. Note that, one output unit may correspond to the split nodes belonging to different trees. Each tree has independent leaf node distribution $\bm{\pi}$ (denoted by distribution histograms and curves in the leaf nodes of the DLDLF and the DRF, respectively). The output of the forest is a mixture of the tree predictions. $\mathbf{f}(\cdot;\bm{\Theta})$ and $\bm{\pi}$ are learned jointly in an end-to-end manner.}\label{fig:DRForest}
\end{figure*}

To jointly learn the deep facial features at split nodes
and the age distributions at leaf nodes in our Deep Differentiable Random Forests, we apply an alternating optimization strategy: first
we fix the leaf nodes parameters and optimize split node parameters
as well as the CNN parameters (feature learning)
by Back-propagation. Then, we fix split node parameters
and optimize the age distributions at leaf nodes by Variational
Bounding~\cite{Ref:Jordan99,Ref:Yuille03}. These two learning steps are alternatively
performed to jointly optimize feature learning and
estimator modeling for age estimation. Additionally, these
two learning steps are non-convex optimization problems
(except for optimizing the age distributions at leaf nodes
in DLDLFs), thus both Gradient Descent and Variational
Bounding require ``good'' parameter initializations to avoid
converging to poor local minimal. To address this problem,
we introduce two Deterministic Annealing (DA) processes~\cite{Ref:Rose98deterministicannealing,Ref:UedaN94,Ref:UedaN98} into these two learning steps, respectively,
which can avoid many poor local optima during optimization
and obtain better estimates of tree parameters free of
initializations. Finally, to learn the ensemble of multiple
trees (forest), we explicitly define the forest loss as the
average of the losses of all the individual trees and allow
the split nodes from different trees to be connected to the
same output unit of the feature learning function. In this
way, the split node parameters of all the individual trees
can be learned jointly. Fig.~\ref{fig:DRForest} illustrates a sketch chart of our DLDLF and DRF, where each forest consists of two trees is shown.

We evaluate our algorithms on three standard datasets for real age estimation methods: MORPH~\cite{Ref:MORPH06}, FG-NET~\cite{Ref:PanisLTC16} and the Cross-Age Celebrity Dataset (CACD)~\cite{Ref:ChenCH15}. Experimental results demonstrate that our algorithms outperform several state-of-the-art methods on these three datasets.


The contributions of this paper are five folds:

1) We propose Deep Label Distribution Learning Forest (DLDLF) and Deep Regression Forest (DRF), two end-to-end models, to deal with inhomogeneous data by jointly learning input-dependent data partition at split nodes and age distribution at leaf nodes.

2) Based on Variational Bounding, the convergences of our update rules for leaf nodes in DLDLFs and DRFs are mathematically guaranteed.

3) We introduce Deterministic Annealing processes into the learning of DLDLFs and DRFs, which can avoid many poor local optima during optimization and obtain better estimates of tree parameters free of initial parameter values.

4) We propose a strategy to learn the ensemble of multiple trees, which is different from~\cite{Ref:Kontschieder15}, but we show it is effective.

5) We apply DLDLFs and DRFs to three standard age estimation benchmarks, and achieve state-of-the-art results.

This paper summarizes two of our preliminary works~\cite{Ref:Shen17,Ref:Shen18} into a unified optimization framework, \emph{i.e.},
alternatively learning split nodes by Back-propagation and learning leaf nodes by Variational Bounding and has following extensions: First, we introduce two methodological improvements, \emph{i.e.},  the two Deterministic Annealing processes
introduced into the learning of split and leaf nodes,
respectively, to avoid poor local optima and obtain better
estimates of tree parameters free of initial parameter values.
Second, we provide more experimental results and discussions,
such as ablation experiments to study the influence of
different designs and variants of our methods and updated
state-of-the-art results on the three age estimation datasets.



%
%
\section{Related Work}\label{sec:related_work}
\textbf{Age Estimation}
One way to tackle precise facial age estimation is to search for a kernel-based global non-linear mapping, like kernel support vector regression \cite{Ref:GuoMFH09} or kernel partial least squares regression \cite{Ref:GuoM11}. The basic idea is to learn a low-dimensional embedding of the aging manifold \cite{Ref:Guo08}. However, global non-linear mapping algorithms may be biased \cite{Ref:Huang17}, due to the inhomogeneous properties of the input data. Another way is to adopt divide-and-conquer approaches, which partition the data space and learn multiple local regressors. But hierarchical regression~\cite{Ref:HanOLJ15} or tree based regression~\cite{Ref:Montillo09} approaches made hard partitions according to ages, which is problematic because the subsets of facial images may not be homogeneous for learning local regressors. Huang \emph{et al.}~\cite{Ref:Huang17} proposed Soft-margin Mixture of Regressions (SMMR) to address this issue, which found homogeneous partitions in the joint input-output space, and learned a local regressor for each partition. But their regression model cannot be integrated with any deep networks as an end-to-end model.

Several researchers formulated age estimation as an ordinal regression problem \cite{Ref:Chang11,Ref:Niu16,Ref:ChenZDLR17}, because the relative order among the age labels is also important information. They trained a series of binary classifiers to partition the samples according to ages, and estimated ages by summing over the classifier outputs. Thus, ordinal regression is limited by its lack of scalability~\cite{Ref:Huang17}. Some other researchers formulated age estimation as a label distribution learning (LDL) problem~\cite{Ref:geng2016label}, which paid attention to modeling the cross-age correlations, based on the observation that faces at close ages look similar. LDL based age estimation methods~\cite{Ref:Geng2010Facial,Ref:Geng13,Ref:Yang16} achieved promising results, but these LDL methods assume that a label distribution should be represented by a maximum entropy model~\cite{Ref:Berger96}, where the exponential part of this model restricts the generality of the distribution form. On the contrary, our method, DLDLF, expresses a label distribution by a linear combination of the label distributions of training data, and thus have no restrictions on the distributions (\emph{e.g.}, no requirement of the maximum entropy model).


With the rapid development of deep networks, more and more end-to-end CNN based age estimation methods~\cite{Ref:Rothe16,Ref:Niu16,Ref:Agustsson17,Ref:Pan18,Ref:Gao2017Deep} have been proposed to address this non-linear regression problem. But how to deal with inhomogeneous data is still an open issue.

\textbf{Random Forests}
Random Forests or randomized decision trees~\cite{Ref:Ho95,Ref:Amit97,Ref:Breiman01,Ref:Shotton13}, are a popular ensemble predictive model suitable for many machine learning tasks, such as supervised learning~\cite{Ref:Breiman01}, semi-supervised learning~\cite{Ref:LeistnerSSB09} and multiple instance learning~\cite{Ref:LeistnerSB10}. Each decision tree consists of several split nodes and leaf nodes. Tree growing is usually based on greedy algorithms which make locally-optimal hard data partition decisions at each split node. Thus, this makes it intractable to integrate decision trees with deep networks in an end-to-end learning manner. Some efforts have been made to combine these two worlds~\cite{Ref:Kontschieder15,Ref:Ioannou16,Ref:LeeGT18}. The newly proposed Deep Neural Decision Forests (DNDFs)~\cite{Ref:Kontschieder15} overcame this problem by introducing a soft differentiable decision function at the split nodes and a global loss function defined on a tree, which ensured that the split node parameters can be learned by back-propagation and leaf node predictions can be updated by a discrete iterative function.

Our methods are inspired by Deep Neural Decision
Forests (DNDFs)~\cite{Ref:Kontschieder15}, but differ in their objectives (label distribution learning/regression \emph{vs} classification).  Extending
differentiable decision trees to deal with label distribution
learning/regression is non-trivial, since there are some technical
difficulties in learning leaf node predictions. Although
a step-size free update function was given in DNDFs to
update leaf node predictions, it was only proved to converge
for a classification loss. Consequently, it is unclear
how to obtain such an update function for other objectives,
especially for regression, since the distribution of the output
space for regression is continuous, but the distribution of the
output space for classification is discrete. We show that the
update functions for both the LDL loss and the regression
loss can be derived from Variational Bounding and the one
given in~\cite{Ref:Kontschieder15} is also a special case of Variational Bounding.
In addition, we introduce two DA processes into the optimization
of our Deep Differentiable Random Forests, which lead to better
estimates of tree parameters free of initial parameter values.
Last but not least, the strategies used in our deep Random
Forests and DNDFs to learn the ensemble of multiple trees
(forests) are different: We explicitly define a loss function for a forest, which allows the split nodes from different trees to
be connected to the same output unit of the feature learning
function (See Fig.~\ref{fig:DRForest}) and enables that all trees in a DLDLF
or a DRF can be learned jointly; while only the loss function
for a single tree is defined in DNDFs, which only allows
trees in a DNDF to be learned alternatively. As shown in
our experiments (Sec.~\ref{sec:param_discussion}), our ensemble strategy can get
better results by using more trees, but by using the ensemble
strategy proposed in DNDFs, the results of forests are even
worse than those for a single tree.

One recent work proposed Neural Regression Forest (NRF)~\cite{Ref:Roy16} for depth estimation, which is similar to our
DRF, but there are two main differences between an NRF
and a DRF. The first difference is all the split nodes in a
DRF are connected to the top layer of a single CNN, but
every split node in an NRF is connected to a distinct CNN.
Therefore, an NRF can be only connected to very shallow
CNNs (as they did in their experiments), otherwise, the
computational cost is extremely high. But, the representation
learning ability of these shallow CNNs is limited. The
second difference is the convergence of our update rule
for leaf nodes is mathematically guaranteed by Variational
Bounding, but the convergence of the update rule for leaf
nodes used in the NRF was not guaranteed.

\section{Differential Decision Trees}\label{sec:ddt}
Since both DLDLFs and DRFs are based on differential decision trees~\cite{Ref:Kontschieder15}, we introduce this tree model first in this section.

\begin{figure}[!t]
\centering
\includegraphics[trim=5cm 5cm 5cm 7cm, clip=true, width=1.0\linewidth]{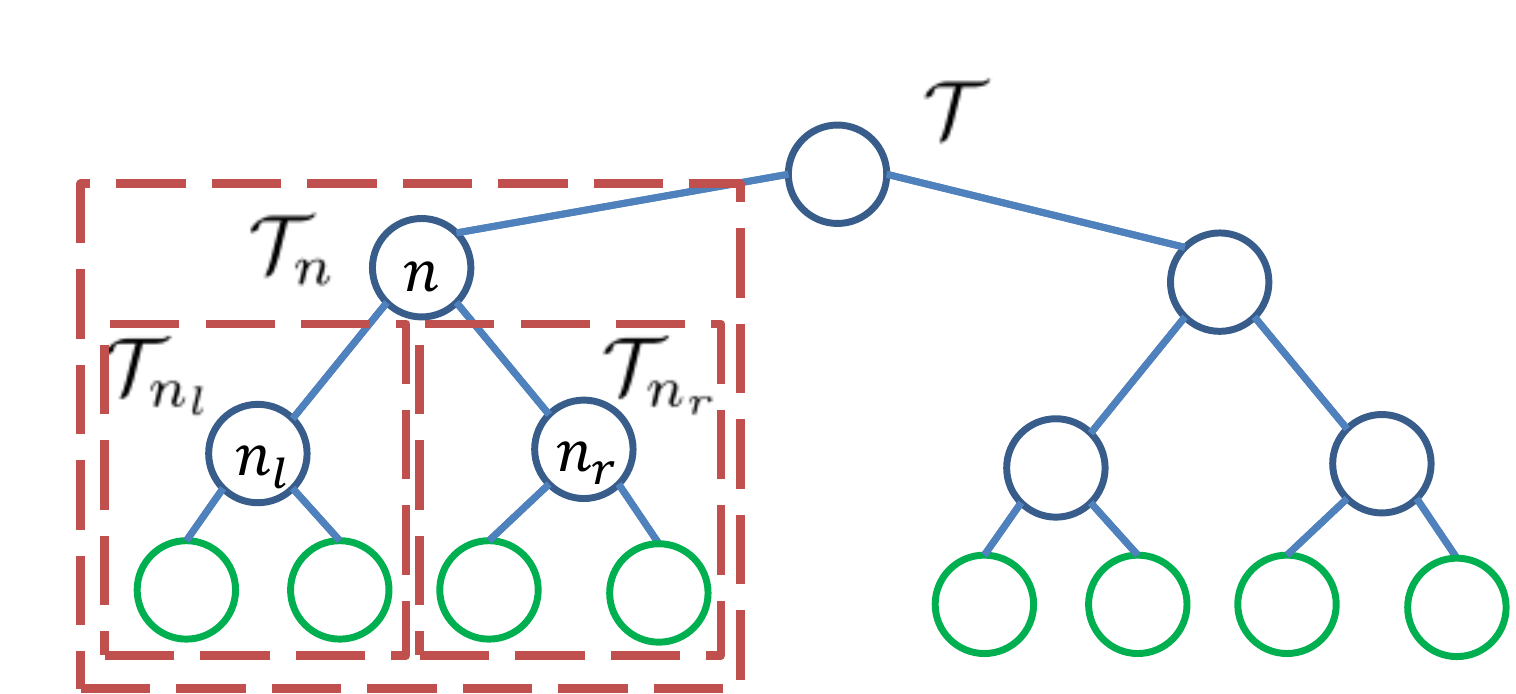}
\caption{The subtree rooted at node $n$: $\mathcal {T}_n$ and its left and right subtrees: $\mathcal {T}_{n_l}$ and $\mathcal {T}_{n_r}$.}\label{fig:Subtree}
\end{figure}

Let $\mathcal {X}$ and $\mathcal {Y}$ denote the input and output spaces, respectively. A differential decision tree $\mathcal{T}$ consists of a set of split nodes $\mathcal {N}$ and a set of leaf nodes $\mathcal {L}$. Each split node $n\in\mathcal {N}$ defines a split function $s_n(\cdot;\bm{\Theta}):\mathcal {X}\rightarrow[0,1]$ parameterized by $\bm{\Theta}$ to determine whether a sample is sent to the left or right subtree. Each leaf node $\ell\in\mathcal {L}$ holds a distribution $\bm{\pi}_{\ell}$ over $\mathcal {Y}$. Following~\cite{Ref:Kontschieder15}, we use a soft split function  $s_n(\mathbf{x};\bm{\Theta}) = \sigma(f_{\varphi(n)}(\mathbf{x};\bm{\Theta}))$, where $\sigma(\cdot)$ is a sigmoid function, $\mathbf{x} \in \mathcal {X}$ and $\mathbf{f}:\mathbf{x}\rightarrow\mathbb{R}^M$ is a real-valued feature learning function depending on the sample $\mathbf{x}$ and the parameter $\bm{\Theta}$. $\mathbf{f}$ can take any forms. In our DLDLFs and DRFs, it is a CNN and $\bm{\Theta}$ is the network parameter. $\varphi(\cdot)$ is an index function to specify the correspondence between
the split nodes and output units of $\mathbf{f}$, which is randomly
assigned before tree learning and then fixed. An example to demonstrate $\varphi(\cdot)$ is shown in Fig.~\ref{fig:DRForest} (There are two trees with index functions in each forest, $\varphi_1$ and $\varphi_2$ for each). Then, the probability of the sample $\mathbf{x}$ falling into leaf node $\ell$ is given by
\begin{equation}\label{eqn:prob_leaf}
P(\ell|\mathbf{x};\bm{\Theta})=\prod_{n\in\mathcal {N}}s_n(\mathbf{x};\bm{\Theta})^{\mathbf{1}(\ell\in\mathcal {L}_{n_l})}(1-s_n(\mathbf{x};\bm{\Theta}))^{\mathbf{1}(\ell\in\mathcal {L}_{n_r})},
\end{equation}
where $\mathbf{1}(\cdot)$ is an indicator function and $\mathcal {L}_{n_l}$ and $\mathcal {L}_{n_r}$ denote the sets of leaf nodes held by the subtrees $\mathcal {T}_{n_l}$, $\mathcal {T}_{n_r}$ rooted at the left and right children $n_l, n_r$ of node $n$ (shown in Fig.~\ref{fig:Subtree}), respectively.

Note that, there are two parameters introduced in this tree model: 1) the split node parameter $\bm{\Theta}$ and 2) the distributions $\bm{\pi}$ held by the leaf nodes. Tree learning requires the estimation of these two parameters. Let $\mathcal {S}$ be a training set and $R(\bm{\pi},\bm{\Theta};\mathcal {S})$ be an objective function for an arbitrary learning task, \emph{e.g.}, classification, label distribution learning or regression, then the best parameters $(\bm{\Theta}^{\ast},\bm{\pi}^{\ast})$ are determined by solving
\begin{equation} \label{eqn:best_para_ldl}
(\bm{\Theta}^{\ast},\bm{\pi}^{\ast}) = \arg\min_{\bm{\Theta},\bm{\pi}}R(\mathbf{\bm{\pi}},\bm{\Theta};\mathcal {S}).
\end{equation}
To solve Eq.~\ref{eqn:best_para_ldl}, we consider an alternating optimization strategy: First, we fix $\bm{\pi}$ and optimize $\bm{\Theta}$ by Back-propagation; Then, we fix $\bm{\Theta}$ and optimize $\bm{\pi}$ by Variational Bounding~\cite{Ref:Jordan99,Ref:Yuille03}. These two learning steps are performed alternatively, until convergence or a maximum number of iterations is reached (as described in the experiments). We show that this optimization framework is unified for learning both DLDLFs and DRFs in Sec.~\ref{sec:DLDLFs} and Sec.~\ref{sec:DRFs}, respectively, as well as for learning the tree models for classification~\cite{Ref:Kontschieder15} in the Appendix.

\section{Age Estimation by Deep Label Distribution Forests} \label{sec:DLDLFs}
In this section, we describe Deep Label Distribution Learning Forests (DLDLFs) for age estimation. Since a forest is an ensemble of decision trees, We first introduce how to learn a single decision tree by label distribution learning, then describe the learning of a forest.
\subsection{Problem Formulation}\label{sec:pro_form_ldl}
\begin{figure}[!th]
\centering
\includegraphics[trim=0cm 0cm 0cm 0cm, clip=true, width=1.0\linewidth]{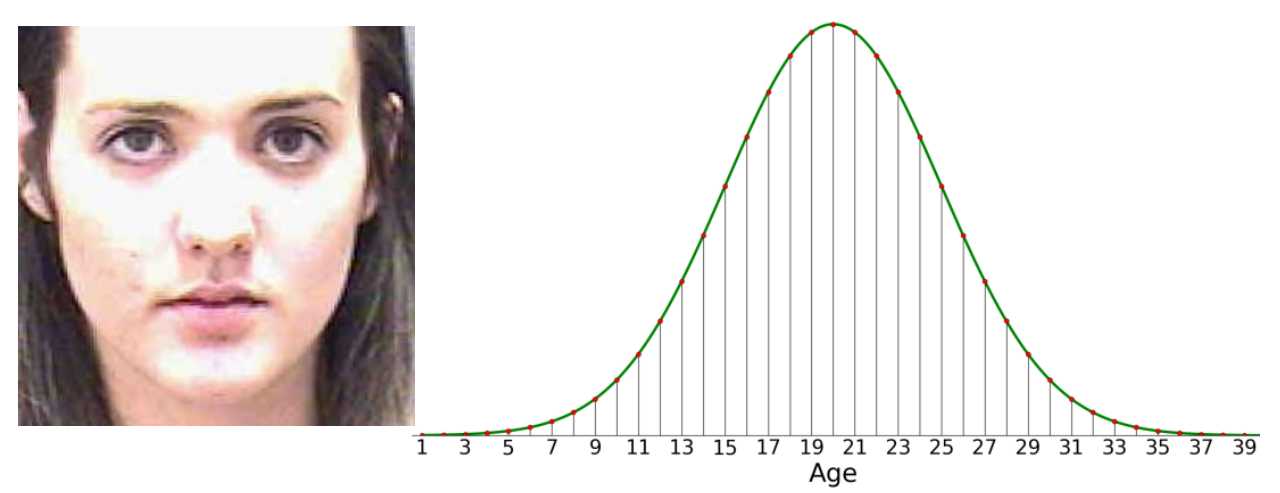}
\caption{Generated label distribution for a facial image at the
chronological age of 20 ($\alpha=20$).}\label{fig:Age_Distribution}
\end{figure}

We formulate age estimation as an LDL problem: Let $\mathcal {X} = \mathbb{R}^m$ denote the input facial image space and $\mathcal {Y}=\{y_1,y_2,\ldots,y_C\}$ denote the complete and order set of age labels, where $C$ is the number of possible age values. For a facial image $\mathbf{x}\in\mathcal {X}$, its chronological age value is $y\in\mathcal {Y}$. To generate a proper label distribution $\mathbf{d}=(d_{\mathbf{x}}^{y_1}, d_{\mathbf{x}}^{y_2},\ldots,d_{\mathbf{x}}^{y_C})^{\top}\in\mathbb{R}^C$, where $d_{\mathbf{x}}^{y_c}\in[0,1]$ and $\sum_{c=1}^Cd_{\mathbf{x}}^{y_c}=1$, for this facial image $\mathbf{x} $, following~\cite{Ref:Geng13,Ref:Gao2017Deep}, we use a Gaussian distribution whose mean is the chronological age $y$:
\begin{align}\label{eqn:age_distribution}
d_{\mathbf{x}}^{y_c}=\frac{p_{\mathcal{N}}(y_c;y,\alpha)}{\sum_{k=1}^Cp_\mathcal{N}(y_k;y,\alpha)},
\end{align}
where $p_\mathcal{N}(y_c;y,\alpha)=\frac{1}{\sqrt{2\pi}\alpha}\exp(-\frac{(y_c-y)^2}{2\alpha^2})$ and $\alpha$ is a pre-defined standard deviation. Fig.~\ref{fig:Age_Distribution} illustrates an example of such a label distribution generated for a facial image at the chronological age of 20 ($\alpha=10$). Observed that, this label distribution explicitly model the cross age correlations, since both the chronological age 20 and the neighboring ages 19 and 21 can be used to describe the appearance of this 20-year-old face, due to the appearance similarity of the neighboring ages.

We have formulated age estimation as an LDL problem, then our goal is to learn a mapping function $\mathbf{g}:\mathbf{x}\rightarrow\mathbf{d}$ between an facial image $\mathbf{x}$ and its corresponding label distribution $\mathbf{d}$ by a decision tree based model $\mathcal {T}$ described in Sec.~\ref{sec:ddt}. Since our target, label distribution, is a discrete distribution, accordingly each leaf node $\ell\in\mathcal {L}$ in the LDL tree $\mathcal {T}$ holds a probability mass distribution $\bm{\pi}_{\ell}=(\pi_{\ell_1},\pi_{\ell_2},\ldots,\pi_{\ell_C})^{\mathrm{T}}$ over $\mathcal {Y}$, \emph{i.e.}, $\pi_{\ell_c}\in[0,1]$ and $\sum_{c=1}^C\pi_{\ell_c}=1$. Then the output of the tree $\mathcal {T}$ w.r.t. $\mathbf{x}$, \emph{i.e.}, the mapping function $g$, is given by
\begin{align}
\mathbf{g}(\mathbf{x};\bm{\Theta},\mathcal {T}) = \sum_{\ell\in\mathcal{L}}P(\ell|\mathbf{x};\bm{\Theta})\bm{\pi}_{\ell}.
\end{align}

\subsection{Tree Optimization}\label{sec:tree_opt_ldl}
Given a training set $\mathcal {S}=\{(\mathbf{x}_i,\mathbf{d}_i)\}_{i=1}^N$, our goal is to learn a LDL tree $\mathcal {T}$ described in in Sec.~\ref{sec:pro_form_ldl} which can output a distribution $\mathbf{g}(\mathbf{x}_i;\bm{\Theta}, \mathcal {T})$ similar to $\mathbf{d}_i$ for each sample $\mathbf{x}_i$. To this end, a straightforward way is to minimize the Kullback-Leibler (K-L) divergence between each $\mathbf{g}(\mathbf{x}_i;\bm{\Theta}, \mathcal {T})$ and $\mathbf{d}_i$, or equivalently to minimize the following cross-entropy loss:
\begin{align}\label{eqn:tree_loss_ldl}
R(\bm{\pi},\bm{\Theta};\mathcal {S})&=-\frac{1}{N}\sum_{i=1}^N\sum_{c=1}^{C}d_{\mathbf{x}_i}^{y_c}\log(g_c(\mathbf{x}_i;\bm{\Theta}, \mathcal {T}))\nonumber\\
&=-\frac{1}{N}\sum_{i=1}^N\sum_{c=1}^{C}d_{\mathbf{x}_i}^{y_c}\log\Big(\sum_{\ell\in\mathcal{L}}P(\ell|\mathbf{x}_i;\bm{\Theta})\pi_{\ell_c}\Big),
\end{align}
where $\bm{\pi}$ denotes the distributions held by all the leaf nodes $\mathcal {L}$ and $g_c(\mathbf{x}_i;\bm{\Theta}, \mathcal {T})$ is the $c$-th output unit of $\mathbf{g}(\mathbf{x}_i;\bm{\Theta}, \mathcal {T})$.

Learning the tree $\mathcal {T}$ requires the estimation of two parameters: 1) the split node parameter $\bm{\Theta}$ and 2) the distributions $\bm{\pi}$ held by the leaf nodes. Next we introduce the optimization process in detail following the framework described in Sec.~\ref{sec:ddt}.

\subsubsection{Learning split nodes w/ Deterministic Annealing by
Gradient Descent}\label{sec:split_node_ldl}
In this section, we describe how to learn the parameter $\bm{\Theta}$ for split nodes, when the distributions held by the leaf nodes $\bm{\pi}$ are fixed.  We found that, empirically (also shown in~\cite{Ref:Kontschieder15}) minimization of $R(\bm{\pi},\bm{\Theta};\mathcal {S})$ w.r.t. $\bm{\Theta}$ would gradually produce almost hard data partitions, \emph{i.e.}, $P(\ell|\mathbf{x}_i;\bm{\Theta})$ approaches 0 or 1. But it is better to enforce $P(\ell|\mathbf{x}_i;\bm{\Theta})$ to be uniform for all $\ell \in \mathcal{L}$, \emph{i.e.}, maintain more uncertainty, at the beginning of minimization. Inspired by ~\cite{Ref:Rose98deterministicannealing}, we introduce a Deterministic Annealing (DA) process into this optimization, which minimizes $R(\bm{\pi},\bm{\Theta};\mathcal {S})$ subject to a specified level of uncertainty. The level of uncertainty is measured by the Shannon entropy:
\begin{align}\label{eqn:entropy}
H(\bm{\Theta};\mathcal{S})=-\frac{1}{N}\sum_{i=1}^N\sum_{\ell\in\mathcal{L}}P(\ell|\mathbf{x}_i;\bm{\Theta})\log P(\ell|\mathbf{x}_i;\bm{\Theta}).
\end{align}
Then we reformulate the original loss function Eq.~\ref{eqn:tree_loss_ldl} to be
\begin{align}\label{eqn:new_tree_loss_ldl}
E(\bm{\pi},\bm{\Theta};\mathcal {S},T)=R(\bm{\pi},\bm{\Theta};\mathcal {S})-TH(\bm{\Theta};\mathcal{S}),
\end{align}
where $T$ is the temperature parameter. During the DA process, $E(\bm{\pi},\bm{\Theta};\mathcal {S},T)$ is then gradually deformed to its original form, \emph{i.e.}, $R(\bm{\pi},\bm{\Theta};\mathcal {S})$, by decreasing the temperature $T\rightarrow 0$. From the DA viewpoint, minimization the original loss function corresponds to a ``zero temperature'' system, where each input sample must make a hard decision about which leaf node it would fall into. This is hard at the beginning of the minimization. On the other hand, starting
at high $T$ smoothes the loss function $E(\bm{\pi},\bm{\Theta};\mathcal {S},T)$ making
it easier to get a good minimum, which can be traced by
slowly decreasing $T$ (``cooling'' the system). By introducing
this DA process, we start with each input sample equally influencing
all leaf nodes and gradually localize the influence.
This gives us some intuition as to how the system searches
for a better optimum~\cite{Ref:Rose98deterministicannealing}. We use a simple cooling schedule to decrease $T$ during optimization: $T\leftarrow \eta T$, where $\eta$ is a constant less than 1.

We compute the gradient of the loss $E(\bm{\pi},\bm{\Theta};\mathcal {S},T)$ w.r.t. $\bm{\Theta}$ by the chain rule:
\begin{align}
\frac{\partial{E}(\bm{\pi},\bm{\Theta};\mathcal {S},T)}{\partial{\bm{\Theta}}}=\sum_{i=1}^N\sum_{n\in\mathcal {N}}\frac{\partial E(\bm{\pi},\bm{\Theta};\mathcal {S},T)}{{\partial}f_{\varphi(n)}(\mathbf{x}_i;\bm{\Theta})}\frac{{\partial}f_{\varphi(n)}(\mathbf{x}_i;\bm{\Theta})}{\partial{\bm{\Theta}}},
\end{align}
where only the first term depends on the tree. The second term depends on the specific type of the function $f_{\varphi(n)}$. The first term is given by
\begin{align} \label{eqn:gradient_ldl}
\frac{\partial{E}(\bm{\pi},\bm{\Theta};\mathcal {S},T)}{{\partial}f_{\varphi(n)}(\mathbf{x}_i;\bm{\Theta})}&=\frac{\partial{R}(\bm{\pi},\bm{\Theta};\mathcal {S})}{{\partial}f_{\varphi(n)}(\mathbf{x}_i;\bm{\Theta})}-T\frac{\partial{H}(\bm{\Theta};\mathcal{S})}{{\partial}f_{\varphi(n)}(\mathbf{x}_i;\bm{\Theta})}\nonumber\\
&=\frac{1}{N}\Big(s_n(\mathbf{x}_i;\bm{\Theta})\big(D^{n_r}_i-TS^{n_r}_i\big)\nonumber\\
&-\big(1-{s}_n(\mathbf{x}_i;\bm{\Theta})\big)\big(D^{n_l}_i-TS^{n_l}_i\big)\Big),
\end{align}
where for a generic node $n\in\mathcal{N}$
\begin{align} \label{eqn:D_inter}
D^{n}_i=\sum_{c=1}^Cd_{\mathbf{x}_i}^{y_c}\frac{g_c(\mathbf{x}_i;\bm{\Theta}, \mathcal {T}_n)}{g_c(\mathbf{x}_i;\bm{\Theta}, \mathcal {T})}=\sum_{c=1}^Cd_{\mathbf{x}_i}^{y_c}\frac{\sum_{\ell\in\mathcal{L}_n}P(\ell|\mathbf{x}_i;\bm{\Theta})\pi_{\ell_c}}{\sum_{\ell\in\mathcal{L}}P(\ell|\mathbf{x}_i;\bm{\Theta})\pi_{\ell_c}},
\end{align}
and
\begin{align} \label{eqn:S_inter}
S^{n}_i=\sum_{\ell\in\mathcal{L}_n}\Big( P(\ell|\mathbf{x}_i;\bm{\Theta}) + P(\ell|\mathbf{x}_i;\bm{\Theta})\log P(\ell|\mathbf{x}_i;\bm{\Theta})\Big).
\end{align}
Both $D^{n}_i$ and $S^{n}_i$ can be efficiently computed for all nodes $n$ in the tree $\mathcal {T}$ by a single pass over the tree. Observing that $D^{n}_i=D_i^{n_l} + D_i^{n_r}$ and $S^{n}_i=S_i^{n_l} + S_i^{n_r}$, the computation for $D^{n}_i$ and $S^{n}_i$ can be started at the leaf nodes and conducted in a bottom-up manner. Following Eq.~\ref{eqn:gradient_ldl}, the split node parameters $\bm{\Theta}$ can be learned by standard Back-propagation.
\subsubsection{Learning leaf nodes by Variational Bounding}
Note that, since the entropy term introduced in Eq.~\ref{eqn:new_tree_loss_ldl} is constant w.r.t. to $\bm{\pi}$, thus it does not influence the learning of leaf nodes. By fixing the parameter $\bm{\Theta}$, we show how to learn the distributions at the leaf nodes $\bm{\pi}$, which is a constrained convex optimization problem:
\begin{align}
\min_{\bm{\pi}}R(\bm{\pi},\bm{\Theta};\mathcal {S}),\textbf{s.t.}, \forall\ell,\sum_{c=1}^C\pi_{\ell_c}=1.
\end{align}
We address this constrained convex optimization problem
by Variational Bounding~\cite{Ref:Jordan99,Ref:Yuille03}. In Variational Bounding,
the original objective function to be minimized gets replaced
by tight upper bounds in an iterative manner. A tight upper bound for the loss function $R(\bm{\pi},\bm{\Theta};\mathcal {S})$ can be obtained by Jensen's inequality:
\begin{align}\label{eqn:jensen_ldl}
R(\bm{\pi},\bm{\Theta};\mathcal {S})=-\frac{1}{N}\sum_{i=1}^N\sum_{c=1}^{C}d_{\mathbf{x}_i}^{y_c}\log\Big(\sum_{\ell\in\mathcal{L}}P(\ell|\mathbf{x}_i;\bm{\Theta})\pi_{\ell_c}\Big)\nonumber\\
\leq -\frac{1}{N}\sum_{i=1}^N\sum_{c=1}^{C}d_{\mathbf{x}_i}^{y_c}\sum_{\ell\in\mathcal{L}}\rho(\ell|{\bar{\pi}_{\ell_c}},\mathbf{x}_i)\log\Big(\frac{P(\ell|\mathbf{x}_i;\bm{\Theta})\pi_{\ell_c}}{\rho(\ell|{\bar{\pi}_{\ell_c}},\mathbf{x}_i)}\Big),
\end{align}
where $\rho(\ell|{{\pi}_{\ell_c}},\mathbf{x}_i)=\frac{P(\ell|\mathbf{x}_i;\bm{\Theta})\pi_{\ell_c}}{g_c(\mathbf{x}_i;\bm{\Theta}, \mathcal {T})}$ and it has the property that $\rho(\ell|{{\pi}_{\ell_c}},\mathbf{x}_i) \in [0, 1]$ and $\sum_{\ell\in\mathcal{L}}\rho(\ell|{{\pi}_{\ell_c}},\mathbf{x}_i)=1$. Note that when $\bm{\pi}=\bar{\bm{\pi}}$, the equality holds, which indicates this upper bound is tight. We define
\begin{align}
\phi(\bm{\pi},\bar{\bm{\pi}}) = -\frac{1}{N}\sum_{i=1}^N\sum_{c=1}^{C}d_{\mathbf{x}_i}^{y_c}\sum_{\ell\in\mathcal{L}}\rho(\ell|{\bar{\pi}_{\ell_c}},\mathbf{x}_i)\log\Big(\frac{P(\ell|\mathbf{x}_i;\bm{\Theta})\pi_{\ell_c}}{\rho(\ell|{\bar{\pi}_{\ell_c}},\mathbf{x}_i)}\Big).
\end{align}
Then $\phi(\bm{\pi},\bar{\bm{\pi}})$ is a tight upper bound for $R(\bm{\pi},\bm{\Theta};\mathcal {S})$, which has the properties that for any $\bm{\pi}$ and $\bar{\bm{\pi}}$, $\phi(\bm{\pi},\bar{\bm{\pi}})\geq \phi(\bm{\pi},\bm{\pi})=R(\bm{\pi},\bm{\Theta};\mathcal {S})$ and $\phi(\bar{\bm{\pi}},\bar{\bm{\pi}})= R(\bar{\bm{\pi}},\bm{\Theta};\mathcal {S})$. These two properties hold the conditions for Variational Bounding.

Assume that we are at a point $\bm{\pi}^{(t)}$ corresponding to the $t$-th iteration, then $\phi(\bm{\pi},\bm{\pi}^{(t)})$ is a tight upper bound for $R(\bm{\pi},\bm{\Theta};\mathcal {S})$. In the next iteration, $\bm{\pi}^{(t+1)}$ is chosen such that $\phi(\bm{\pi}^{(t+1)},\bm{\pi}^{(t)})\leq R(\bm{\pi}^{(t)},\bm{\Theta};\mathcal {S})$, which implies $R(\bm{\pi}^{(t+1)},\bm{\Theta};\mathcal {S})\leq R(\bm{\pi}^{(t)},\bm{\Theta};\mathcal {S})$. Consequently, we can minimize  $\phi(\bm{\pi},\bar{\bm{\pi}})$ instead of $R(\bm{\pi},\bm{\Theta};\mathcal {S})$ after ensuring that
$R(\bm{\pi}^{(t)},\bm{\Theta};\mathcal {S})=\phi(\bm{\pi}^{(t)},\bar{\bm{\pi}})$, \emph{i.e.}, $\bar{\bm{\pi}} = \bm{\pi}^{(t)}$. So we have
\begin{align}
\bm{\pi}^{(t+1)}=\arg\min_{\bm{\pi}}\phi(\bm{\pi},\bm{\pi}^{(t)}), \textbf{s.t.}, \forall\ell,\sum_{c=1}^C\pi_{\ell_c}=1,
\end{align}
which leads to minimizing the Lagrangian defined by
\begin{align}
\varphi(\bm{\pi},\bm{\pi}^{(t)})=\phi(\bm{\pi},\bm{\pi}^{(t)})+\sum_{\ell\in\mathcal {L}}\lambda_{\ell}(\sum_{c=1}^C\pi_{\ell_c}-1),
\end{align}
where $\lambda_{\ell}$ is the Lagrange multiplier. By setting $\frac{\partial\varphi(\bm{\pi},\bm{\pi}^{(t)})}{\partial{\pi_{\ell_c}}}=0$, we have
\begin{align}\label{eqn:update_leaf_ldl}
\lambda_{\ell} &= \frac{1}{N}\sum_{i=1}^N\sum_{c=1}^Cd_{\mathbf{x}_i}^{y_c}\rho({\ell}|\pi^{(t)}_{\ell_c},\mathbf{x}_i),\nonumber \\
\pi^{(t+1)}_{\ell_c}&=\frac{\sum_{i=1}^Nd_{\mathbf{x}_i}^{y_c}\rho({\ell}|\pi^{(t)}_{\ell_c},\mathbf{x}_i)}{\sum_{c=1}^C\sum_{i=1}^Nd_{\mathbf{x}_i}^{y_c}\rho({\ell}|\pi^{(t)}_{\ell_c},\mathbf{x}_i)}.
\end{align}
Note that, $\pi^{(t+1)}_{\ell_c}$ satisfies that $\pi^{(t+1)}_{\ell_c} \in [0,1]$ and $\sum_{c=1}^C\pi^{(t+1)}_{\ell_c}=1$. Eq.~\ref{eqn:update_leaf_ldl} is the update scheme for distributions held by the leaf nodes. The starting point $\bm{\pi}^{(0)}_{\ell}$ can be simply initialized by the uniform distribution: ${\pi}^{(0)}_{\ell_c} = \frac{1}{C}$.

\subsection{Learning an LDL Forest}\label{sec:forest_learning_ldl}
An LDL forest is an ensemble of LDL decision trees $\mathcal {F}=\{\mathcal {T}^1,\ldots,\mathcal {T}^K\}$. In the training stage, all trees in the forest $\mathcal {F}$ use the same parameters $\bm{\Theta}$ for feature learning function $\mathbf{f}(\cdot;\bm{\Theta})$ (but correspond to different output units of $\mathbf{f}$ assigned by $\varphi$, see Fig.~\ref{fig:DRForest}), but each tree has independent leaf node predictions $\bm{\pi}$. The loss function for a forest is given by averaging the loss functions for all individual trees: 
\begin{align}\label{eqn:forest_loss}
E_{\mathcal {F}}=\frac{1}{K}\sum_{k=1}^K E_{\mathcal{T}^k},
\end{align}
where $E_{\mathcal{T}^k}$ is the loss function for tree $\mathcal{T}^k$ defined by Eq.~\ref{eqn:new_tree_loss_ldl}. To learn $\bm{\Theta}$ by fixing the leaf node predictions $\bm{\pi}$ of all the trees in the forest $\mathcal {F}$, based on the derivation in Sec.~\ref{sec:tree_opt_ldl} and referring to Fig.~\ref{fig:DRForest}, we have
\begin{align} \label{eqn:gradient_forest}
\frac{\partial{E}_{\mathcal {F}}}{\partial{\bm{\Theta}}}=\frac{1}{K}\sum_{i=1}^N\sum_{k=1}^K\sum_{n\in\mathcal {N}_k}\frac{\partial{E}_{\mathcal{T}^k}}{{\partial}f_{\varphi_k(n)}(\mathbf{x}_i;\bm{\Theta})}\frac{{\partial}f_{\varphi_k(n)}(\mathbf{x}_i;\bm{\Theta})}{\partial{\bm{\Theta}}},
\end{align}
where $\mathcal {N}_k$ and $\varphi_k(\cdot)$ are the split node set and the index function of $\mathcal{T}^k$, respectively, and the first term of the right side $\frac{\partial{E}_{\mathcal{T}^k}}{{\partial}f_{\varphi_k(n)}(\mathbf{x}_i;\bm{\Theta})}$ is computed by Eq.~\ref{eqn:gradient_ldl}. Note that, the index function $\varphi_k(\cdot)$ for each tree is randomly assigned before tree learning, and thus split nodes correspond to a subset of output units of $\mathbf{f}$. This strategy is similar to the random subspace method~\cite{Ref:Ho98}, which increases the randomness in training to reduce the risk of over-fitting. For the temperature parameter $T$ introduced in each ${E}_{\mathcal{T}^k}$, we initialize it as a large value $T_0$ ($T_0 > 0$) and decrease it by the simple cooling schedule described in Sec.~\ref{sec:split_node_ldl}: $T \leftarrow \eta T$, where $\eta$ is a constant cooling factor less than 1.

As for $\bm{\pi}$, since each tree in the forest $\mathcal {F}$ has its own leaf node predictions $\bm{\pi}$, we can update them independently by Eq.~\ref{eqn:update_leaf_ldl}.
For implementational convenience, we do not conduct this update scheme on the whole dataset $\mathcal{S}$ but on a set of mini-batches $\mathcal{B}$. The training procedure of an LDLF is shown in Algorithm.~\ref{fig:algorithm}.

\begin{algorithm}
\caption{The training procedure of a DLDLF.}
\label{fig:algorithm}
\begin{algorithmic}
\REQUIRE $\mathcal{S}$: training set
\REQUIRE $n_B$: the number of mini-batches to update $\bm{\pi}$,
\REQUIRE $T_0$: a starting temperature parameter
\REQUIRE $\eta$: a constant cooling factor
\STATE Initialize $\bm{\Theta}$ randomly and $\bm{\pi}$ uniformly
\STATE Set $\mathcal{B}=\{\emptyset\}$ and $T=T_0$
\WHILE {Not converge}
\WHILE {$|\mathcal{B}|<n_B$}
\STATE Randomly select a mini-batch $B$ from $\mathcal{S}$
\STATE Update $\bm{\Theta}$ by Gradient Descent (Eq.~\ref{eqn:gradient_forest}, Eq.~\ref{eqn:gradient_ldl}) on $B$
\STATE $\mathcal{B}=\mathcal{B}\bigcup{B}$
\ENDWHILE
\STATE Update $\bm{\pi}$ by iterating Eq.~\ref{eqn:update_leaf_ldl} on $\mathcal{B}$
\STATE $\mathcal{B}=\{\emptyset\}$
\IF {$T\geq 0$}
\STATE {$T \leftarrow \eta T$}
\ENDIF
\ENDWHILE
\end{algorithmic}
\end{algorithm}

In the testing stage, the output of the forest $\mathcal {F}$ is given by averaging the predictions from all the individual trees:
\begin{align}
\mathbf{g}(\mathbf{x};\bm{\Theta},\mathcal {F})=\frac{1}{K}\sum_{k=1}^K\mathbf{g}(\mathbf{x};\bm{\Theta},\mathcal {T}^k).
\end{align}
Then the predicted age value is given by $\hat{y} = y_{c^{\ast}}$, where $ c^{\ast} = \arg\min_cg_c(\mathbf{x};\bm{\Theta}, \mathcal {F})$.
\section{Age Estimation by Deep Regression Forests} \label{sec:DRFs}
In this section, we describe Deep Regression Forests (DRFs) for age estimation. Similar to the previous section, we first introduce how to learn a single differentiable regression tree, then describe how to learn tree ensembles to form a forest.
\subsection{Problem Formulation}
We formulate age estimation as a regression problem, where we regard age as a continues numerical value: Let $\mathcal {X} = \mathbb{R}^m$ denote the input facial image space and $\mathcal {Y}=\mathbb{R}$ denote the output age space. For a facial image $\mathbf{x}\in\mathcal {X}$, its chronological age value is $y\in\mathcal {Y}$. The objective of regression is to find a mapping function ${g}:\mathbf{x}\rightarrow{y}$ between an input sample $\mathbf{x}$ and its output target ${y}$. A standard way to address this problem is to model the conditional probability function $p({y}|\mathbf{x})$, so that the mapping is given by
\begin{align}
\hat{{y}} = \mathbf{g}(\mathbf{x}) = \int{y}p({y}|\mathbf{x})d{y}.
\end{align}

We propose to model this conditional probability by a decision tree based structure $\mathcal {T}$ described in Sec.~\ref{sec:ddt}. Each leaf node $\ell\in\mathcal {L}$ in the regression tree $\mathcal {T}$ holds a probability density distribution $\pi_{\ell}({y})$ over $\mathcal {Y}$, i.e, $\int\pi_{\ell}({y})d{y}=1$. The conditional probability function $p({y}|\mathbf{x};\mathcal {T})$ given by the tree $\mathcal {T}$ is
\begin{align}
p({y}|\mathbf{x};\mathcal {T})=\sum_{\ell\in\mathcal{L}}P(\ell|\mathbf{x};\bm{\Theta})\pi_{\ell}({y}).
\end{align}
Then the mapping between $\mathbf{x}$ and ${y}$ modeled by tree $\mathcal {T}$ is given by
$\hat{{y}} = \mathbf{g}(\mathbf{x};\mathcal {T}) = \int{y}p({y}|\mathbf{x};\mathcal {T})d{y}$.

\subsection{Tree Optimization}\label{sec:tree_opt_reg}
Given a training set $\mathcal {S}=\{(\mathbf{x}_i,{y}_i)\}_{i=1}^N$, learning a regression tree $\mathcal {T}$ leads to minimizing the following negative log likelihood loss:
\begin{align}\label{eqn:tree_loss_reg}
R(\bm{\pi},\bm{\Theta};\mathcal{S})=-\frac{1}{N}\sum_{i=1}^{N}\log(p({y}_{i}|\mathbf{x}_{i}, \mathcal {T}))\nonumber\\
=-\frac{1}{N}\sum_{i=1}^{N}\log\big(\sum_{\ell\in\mathcal{L}}P(\ell|\mathbf{x}_i;\bm{\Theta})\pi_{\ell}({y}_i)\big),
\end{align}
where $\bm{\pi}$ denotes the density distributions contained by all the leaf nodes $\mathcal {L}$. To optimize $R(\bm{\pi},\bm{\Theta};\mathcal{S})$ w.r.t. the split node
parameter $\bm{\Theta}$ and the density distributions $\bm{\pi}$ held by leaf nodes, we also follow the optimization framework described in Sec.~\ref{sec:ddt}, \emph{i.e.}, alternating the following two steps: (1) fixing $\bm{\pi}$ and optimizing $\bm{\Theta}$; (2) fixing $\bm{\Theta}$ and optimizing $\bm{\pi}$, until convergence or a maximum number of iterations is reached.

\subsubsection{Learning split nodes w/ Deterministic Annealing by
Gradient Descent}\label{sec:split_node_reg}
Now, we discuss how to learn the parameter $\bm{\Theta}$ for split nodes, when the density distributions held by the leaf nodes $\bm{\pi}$ are fixed. We introduce the same DA process described in  Sec.~\ref{sec:split_node_ldl} into the optimization for split node parameter $\bm{\Theta}$, which reformulates the original regression loss Eq.~\ref{eqn:tree_loss_reg} as the same form as Eq.~\ref{eqn:new_tree_loss_ldl}, \emph{i.e.}, $E(\bm{\pi},\bm{\Theta};\mathcal {S},T)=R(\bm{\pi},\bm{\Theta};\mathcal {S})-TH(\bm{\Theta};\mathcal{S})$. We use the same cooling schedule in Sec.~\ref{sec:split_node_ldl} to decrease $T$ during optimization: $T \leftarrow \eta T$. Similarly, we
compute the gradient $\frac{\partial{E}(\bm{\pi},\bm{\Theta};\mathcal {S},T)}{\partial{\bm{\Theta}}}$ by the chain rule, and we have
\begin{align} \label{eqn:gradient_reg}
\frac{\partial{E}(\mathbf{\bm{\pi}},\bm{\Theta};\mathcal {S},T)}{{\partial}f_{\varphi(n)}(\mathbf{x}_i;\bm{\Theta})}&=\frac{\partial{R}(\mathbf{\bm{\pi}},\bm{\Theta};\mathcal {S})}{{\partial}f_{\varphi(n)}(\mathbf{x}_i;\bm{\Theta})}-T\frac{\partial{H}(\bm{\Theta};\mathcal{S})}{{\partial}f_{\varphi(n)}(\mathbf{x}_i;\bm{\Theta})}\nonumber\\
&=\frac{1}{N}\Big(s_n(\mathbf{x}_i;\bm{\Theta})\big(\Gamma^{n_r}_i-TS^{n_r}_i\big)\nonumber\\
&-\big(1-{s}_n(\mathbf{x}_i;\bm{\Theta})\big)\big(\Gamma^{n_l}_i-TS^{n_l}_i\big)\Big),
\end{align}
where for a generic node $n\in\mathcal{N}$
\begin{align}\label{eqn:Gamma_inter}
\Gamma_i^{n}=\frac{p({y}_i|\mathbf{x}_i;\mathcal {T}_n)}{p({y}_i|\mathbf{x}_i;\mathcal {T})}=\frac{\sum_{\ell\in\mathcal{L}_n}P(\ell|\mathbf{x}_i;\bm{\Theta})\pi_{\ell}({y}_i)}{p({y}_i|\mathbf{x}_i;\mathcal {T})},
\end{align}
and $S^{n}_i$ is computed by Eq.~\ref{eqn:S_inter}.
$\Gamma^n_{i}$ can be also efficiently computed for all nodes $n$ in the tree $\mathcal {T}$ by a single pass over the tree. Observing that $\Gamma^i_{n}=\Gamma^i_{n_l} + \Gamma^i_{n_r}$, the computation for $\Gamma^i_{n}$ can be started at the leaf nodes and conducted in a bottom-up manner. Based on Eq.~\ref{eqn:gradient_reg}, the split node parameters $\bm{\Theta}$ can be learned by standard Back-propagation.
\subsubsection{Learning leaf nodes by Variational Bounding}
By fixing the split node parameters $\bm{\Theta}$, learning the leaf nodes parameters $\bm{\pi}$ becomes a constrained optimization problem:
\begin{equation}\label{eqn:best_pi_reg}
\min_{\bm{\pi}}R(\bm{\pi},\bm{\Theta};\mathcal {S}),\textbf{s.t.}, \forall\ell,\int\pi_{\ell}({y})d{y}=1.
\end{equation}
For efficient computation, we represent each density distribution $\pi_{\ell}({y})$ by a parametric model. Since ideally each leaf node corresponds to a compact homogeneous subset, we assume that the density distribution $\pi_{\ell}({y})$
in each leaf node is a Gaussian distribution, \emph{i.e.},
\begin{equation}
\pi_{\ell}({y})=\frac{1}{\sqrt{2\pi}{\sigma}_{\ell}}\exp\Big(-\frac{(y-\mu_{\ell})^2}{2\sigma_{\ell}^2}\Big),
\end{equation}
where ${\mu}_{\ell}$ and ${\sigma}_{\ell}$ are the mean and the covariance matrix of the Gaussian distribution. Based on this assumption, Eq.~\ref{eqn:best_pi_reg} is equivalent to minimizing $R(\bm{\pi},\bm{\Theta};\mathcal {S})$ w.r.t. ${\mu}_{\ell}$ and ${\sigma}_{\ell}$. We also propose to address this optimization problem by Variational Bounding~\cite{Ref:Jordan99,Ref:Yuille03}.  To obtain a tight upper bound of $R(\bm{\pi},\bm{\Theta};\mathcal {S})$, we apply Jensen's inequality to it:
\begin{align}\label{eqn:jensen_reg}
R(\bm{\pi},\bm{\Theta};\mathcal{S})=-\frac{1}{N}\sum_{i=1}^{N}\log\big(\sum_{\ell\in\mathcal{L}}P(\ell|\mathbf{x}_i;\bm{\Theta})\pi_{\ell}({y}_i)\big)\nonumber\\
=-\frac{1}{N}\sum_{i=1}^{N}\log\Big(\sum_{\ell\in\mathcal{L}}\rho(\ell|\bar{\bm{\pi}},{y}_i,\mathbf{x}_i)\frac{P(\ell|\mathbf{x}_i;\bm{\Theta})\pi_{\ell}({y}_i)}{\rho(\ell|\bar{\bm{\pi}},{y}_i,\mathbf{x}_i)}\Big)\nonumber\\
\leq  -\frac{1}{N}\sum_{i=1}^{N}\sum_{\ell\in\mathcal{L}}\rho(\ell|\bar{\bm{\pi}},{y}_i,\mathbf{x}_i)\log\Big(\frac{P(\ell|\mathbf{x}_i;\bm{\Theta})\pi_{\ell}({y}_i)}{\rho(\ell|\bar{\bm{\pi}},{y}_i,\mathbf{x}_i)}\Big)\nonumber\\
=R(\bar{\bm{\pi}},\bm{\Theta};\mathcal{S})-\frac{1}{N}\sum_{i=1}^{N}\sum_{\ell\in\mathcal{L}}\rho(\ell|\bar{\bm{\pi}},{y}_i,\mathbf{x}_i)\log\Big(\frac{\pi_{\ell}({y}_i)}{\bar{\pi}_{\ell}({y}_i)} \Big),
\end{align}
where $\rho(\ell|{\bm{\pi}},{y}_i,\mathbf{x}_i) = \frac{P(\ell|\mathbf{x}_i;\bm{\Theta})\pi_{\ell}({y}_i)}{p({y}_i|\mathbf{x}_i;\mathcal {T})}$ and it has the property that $\rho(\ell|{\bm{\pi}},{y}_i,\mathbf{x}_i) \in [0, 1]$ and $\sum_{\ell\in\mathcal{L}}\rho(\ell|{\bm{\pi}},{y}_i,\mathbf{x}_i)=1$. Note that, when $\bm{\pi}=\bar{\bm{\pi}}$, the equality holds, which indicates this
upper bound is tight. Let us define
\begin{align}\label{eqn:phi_reg}
\phi(\bm{\pi},\bar{\bm{\pi}}) = R(\bar{\bm{\pi}},\bm{\Theta};\mathcal{S})-\frac{1}{N}\sum_{i=1}^{N}\sum_{\ell\in\mathcal{L}}\rho(\ell|\bar{\bm{\pi}},{y}_i,\mathbf{x}_i)\log\Big(\frac{\pi_{\ell}({y}_i)}{\bar{\pi}_{\ell}({y}_i)} \Big).
\end{align}
Then $\phi(\bm{\pi},\bar{\bm{\pi}})$ is a tight upper bound for $R(\bm{\pi},\bm{\Theta};\mathcal {S})$, which has the properties that for any $\bm{\pi}$ and $\bar{\bm{\pi}}$, $\phi(\bm{\pi},\bar{\bm{\pi}})\geq \phi(\bm{\pi},\bm{\pi})=R(\bm{\pi},\bm{\Theta};\mathcal {S})$ and $\phi(\bar{\bm{\pi}},\bar{\bm{\pi}})= R(\bar{\bm{\pi}},\bm{\Theta};\mathcal {S})$. These two properties give the conditions for Variational Bounding.

Recall that we parameterize $\pi_{\ell}(\mathbf{y})$ by two parameters: the mean ${\mu}_{\ell}$ and the covariance matrix ${\sigma}_{\ell}$. Let $\bm{\mu}$ and $\bm{\sigma}$ denote these two parameters held by all the leaf nodes $\mathcal{L}$. We define $\psi(\bm{\mu},\bar{\bm{\mu}}) = \phi(\bm{\pi},\bar{\bm{\pi}})$, then $\psi(\bm{\mu},\bar{\bm{\mu}}) \geq \phi(\bm{\pi},\bm{\pi}) = \psi(\bm{\mu},\bm{\mu}) = R(\bm{\pi},\bm{\Theta};\mathcal {S})$, which indicates that $\psi(\bm{\mu},\bar{\bm{\mu}})$ is also a tight upper bound for $R(\bm{\pi},\bm{\Theta};\mathcal {S})$. Assume that we are at a point $\bm{\mu}^{(t)}$ corresponding to the $t$-th iteration, then $\psi(\bm{\mu},\bm{\mu}^{(t)})$ is a tight upper bound for $R(\bm{\pi},\bm{\Theta};\mathcal {S})$. In the next iteration, $\bm{\mu}^{(t+1)}$ is chosen such that $\psi(\bm{\mu}^{(t+1)},\bm{\mu})\leq R(\bm{\pi}^{(t)},\bm{\Theta};\mathcal {S})$, which implies $R(\bm{\pi}^{(t+1)},\bm{\Theta};\mathcal {S})\leq R(\bm{\pi}^{(t)},\bm{\Theta};\mathcal {S})$. Therefore, we can minimize $\psi(\bm{\mu},\bar{\bm{\mu}})$ instead of $R(\bm{\pi},\bm{\Theta};\mathcal {S})$ after ensuring that
$R(\bm{\pi}^{(t)},\bm{\Theta};\mathcal {S})=\psi(\bm{\mu}^{(t)},\bar{\bm{\mu}})$, \emph{i.e.}, $\bar{\bm{\mu}} = \bm{\mu}^{(t)}$. Thus, we have
\begin{align}
\bm{\mu}^{(t+1)} = \arg\min_{\bm{\mu}}\psi(\bm{\mu}, \bm{\mu}^{(t)}).
\end{align}
The partial derivative of $\psi(\bm{\mu}, \bm{\mu}^{(t)})$ w.r.t. ${\mu}_{\ell}$ is computed by
\begin{align}
&\frac{\partial\psi(\bm{\mu}, \bm{\mu}^{(t)})}{\partial{\mu}_{\ell}} = \frac{\partial\phi(\bm{\pi}, \bm{\pi}^{(t)})}{\partial\bm{\mu}_{\ell}}\nonumber\\
&=-\frac{1}{N}\sum_{i=1}^{N}\rho(\ell|\bm{\pi}^{(t)},{y}_i,\mathbf{x}_i){\sigma}^{-1}_{\ell}({y}_i - {\mu}_{\ell}).
\end{align}
By setting $\frac{\partial\psi(\bm{\mu}, \bm{\mu}^{(t)})}{\partial{\mu}_{\ell}} = 0$, we have
\begin{align}\label{eqn:mean}
{\mu}_{\ell}^{(t+1)}=\frac{\sum_{i=1}^N\rho(\ell|\bm{\pi}^{(t)},{y}_i,\mathbf{x}_i){y}_i}{\sum_{i=1}^N\rho(\ell|\bm{\pi}^{(t)},{y}_i,\mathbf{x}_i)}.
\end{align}
Similarly, we define $\nu(\bm{\sigma}, \bar{\bm{\sigma}})=\phi(\bm{\pi},\bar{\bm{\pi}})$, then
\begin{align}
\bm{\sigma}^{(t+1)} = \arg\min_{\bm{\sigma}}\nu(\bm{\sigma}, \bm{\sigma}^{(t)}).
\end{align}
The partial derivative of $\nu(\bm{\sigma}, \bm{\sigma}^{(t)})$ w.r.t. ${\sigma}_{\ell}$ is obtained by
\begin{align}
&\frac{\partial\nu(\bm{\sigma}, \bm{\sigma}^{(t)})}{\partial{\sigma}_{\ell}} = \frac{\partial\phi(\bm{\pi}, \bm{\pi}^{(t)})}{\partial{\sigma}_{\ell}}\nonumber\\
&=-\frac{1}{N}\sum_{i=1}^{N}\rho(\ell|\bm{\pi}^{(t)},{y}_i,\mathbf{x}_i)\big[-\frac{1}{2}{\sigma}_{\ell}^{-1}+\frac{1}{2}{\sigma}_{\ell}^{-2}({y}_i-{\mu}_{\ell}^{(t+1)})^2\big]
\end{align}
By Setting $\frac{\partial\nu(\bm{\sigma}, \bm{\sigma}^{(t)})}{\partial{\sigma}_{\ell}} = 0$, we have
\begin{align} \label{eqn:covar}
{\sigma}_{\ell}^{(t+1)}=\frac{\sum_{i=1}^{N}\rho(\ell|\bm{\pi}^{(t)},{y}_i,\mathbf{x}_i)({y}_i-{\mu}_{\ell}^{(t+1)})^2}{\sum_{i=1}^{N}\rho(\ell|\bm{\pi}^{(t)},{y}_i,\mathbf{x}_i)}.
\end{align}
Eq.~\ref{eqn:mean} and Eq.~\ref{eqn:covar} are the update functions for the density distribution $\bm{\pi}$ held by all leaf nodes, which are step-size free and fast-converged.
\subsubsection{Learning leaf nodes w/ Deterministic Annealing w/o
initialization}
One issue remained is how to initialize the starting point $\bm{\mu}^{(0)}$ and $\bm{\sigma}^{(0)}$. Note that $R(\bm{\pi},\bm{\Theta};\mathcal{S})$ is convex w.r.t. $\bm{\pi}$, but is non-convex w.r.t. $\bm{\mu}$ and $\bm{\sigma}$. Consequently, based on the update functions Eq.~\ref{eqn:mean} and Eq.~\ref{eqn:covar}, $R(\bm{\pi},\bm{\Theta};\mathcal{S})$ may converge to a poor local minimum, if $\bm{\mu}^{(0)}$ and $\bm{\sigma}^{(0)}$ are not well initialized. In our previous work~\cite{Ref:Shen18}, we did k-means clustering on $\{{y}_i\}_{i=1}^N$ to obtain $|\mathcal{L}|$ subsets, then initialized $\bm{\mu}^{(0)}$ and $\bm{\sigma}^{(0)}$ according to cluster assignment. Here, inspired by~\cite{Ref:UedaN94,Ref:UedaN98} we propose a deterministic annealing algorithm for the above optimization problem, which leads to an initialization free solution to avoid poor local minimum. In~\cite{Ref:UedaN94,Ref:UedaN98}, a deterministic annealing Expectation-Maximization (EM) algorithm was presented for Maximum Likelihood Estimation (MLE) problems to obtain better estimates free of the initial parameter values, in which a new posterior parameterized by ``temperature'' is derived by using the principle of maximum entropy and is used for controlling the annealing process. To apply this strategy to our optimization problem, we first rewrite Eq.~\ref{eqn:jensen_reg} in the form of Neal and Hinton's free energy~\cite{Ref:Neal99}:
\begin{align}\label{eqn:free_energy}
&J(\rho,\bm{\pi})=\nonumber\\
&-\frac{1}{N}\sum_{i=1}^N\Big(\mathbb{E}_{\rho}\big[\log\big(P(\ell|\mathbf{x}_i;\bm{\Theta})\pi_{\ell}({y}_i)\big)\big]-\mathbb{E}_{\rho}[\log\rho]\Big),
\end{align}
where $\mathbb{E}_{\rho}[\cdot]$ denotes the expectation w.r.t. conditional probability $\rho(\ell|{\bm{\pi}},{y}_i,\mathbf{x}_i)$. For a fixed $\bm{\pi}$, when $\rho(\ell|{\bm{\pi}},{y}_i,\mathbf{x}_i) = \frac{P(\ell|\mathbf{x}_i;\bm{\Theta})\pi_{\ell}({y}_i)}{p({y}_i|\mathbf{x}_i;\mathcal {T})}$, $J(\rho,\bm{\pi})$ achieves its minimum, \emph{i.e.}, $J(\rho,\bm{\pi}) \equiv R(\bm{\pi},\bm{\Theta};\mathcal{S})$. $\rho(\ell|{\bm{\pi}},{y}_i,\mathbf{x}_i)$ is the posterior in this MLE problem, which plays an important role in the optimization (see Eq.~\ref{eqn:mean} and Eq.~\ref{eqn:covar}). However, since initial values $\bm{\mu}^{(0)}$ and $\bm{\sigma}^{(0)}$ are not guaranteed to be near the true ones, $\rho(\ell|{\bm{\pi}},{y}_i,\mathbf{x}_i)$ may be unreliable at early stages of optimization. Ideally, the influence of this conditional probability should be weakened at the beginning, and as the optimization proceeds, the effect should be strengthened. To this ends, we want to seek another conditional probability $\varrho(\ell|{\bm{\pi}},{y}_i,\mathbf{x}_i)$ to replace $\rho(\ell|{\bm{\pi}},{y}_i,\mathbf{x}_i)$ by extending Eq.~\ref{eqn:free_energy} to a deterministic annealing variant:
\begin{align}\label{eqn:new_free_energy}
&J'(\varrho, \bm{\pi},\tau)=\nonumber\\&-\frac{1}{N}\sum_{i=1}^N\Big(\mathbb{E}_{\varrho}\big[\log\big(P(\ell|\mathbf{x}_i;\bm{\Theta})\pi_{\ell}({y}_i)\big)\big]-\frac{1}{\tau}\mathbb{E}_{\varrho}[\log\varrho]\Big),
\end{align}
where $\frac{1}{\tau}$ is the temperature parameter. Note that, when $\tau =1$, $J'\equiv J$. According to Neal and Hinton's theory~\cite{Ref:Neal99}, the minimization of $J'(\varrho, \bm{\pi},\tau)$ can be performed by the following Coordinate Descent iterations:
\begin{itemize}
\item Set $\varrho^{(t+1)}$ to $\varrho$ that minimizes $J'(\varrho, \bm{\pi}^{(t)},\tau)$
\item Set $\bm{\pi}^{(t+1)}$ to $\bm{\pi}$ that minimizes $J'(\varrho^{(t+1)}, \bm{\pi},\tau)$
\end{itemize}

Given $\bm{\pi}^{(t)}$, $\varrho^{(t+1)}$ is obtained by minimizing $J'(\varrho, \bm{\pi}^{(t)},\tau)$ w.r.t. $\varrho$ under the constraint $\sum_{\ell\in\mathcal{L}}\varrho=1$:
\begin{align}
&\frac{\partial J'(\varrho, \bm{\pi}^{(t)},\tau)}{\partial \varrho} =\nonumber\\
&-\log\big(P(\ell|\mathbf{x}_i;\bm{\Theta})\pi^{(t)}_{\ell}({y}_i)\big)+\frac{1}{\tau}(\log\varrho+1)+\lambda=0,
\end{align}
where $\lambda$ is a Lagrange multiplier. Thus, we have
\begin{align}\label{eqn:new_posterior}
\varrho^{(t+1)}(\ell|{\bm{\pi}^{(t)}},{y}_i,\mathbf{x}_i) = \frac{\big(P(\ell|\mathbf{x}_i;\bm{\Theta})\pi^{(t)}_{\ell}({y}_i)\big)^{\tau}}{\sum_{\ell \in \mathcal{L}}\big(P(\ell|\mathbf{x}_i;\bm{\Theta})\pi^{(t)}_{\ell}({y}_i)\big)^{\tau}}.
\end{align}
Then, by fixing $\varrho(\ell|{\bm{\pi}^{(t)}},{y}_i,\mathbf{x}_i)=\varrho^{(t+1)}(\ell|{\bm{\pi}^{(t)}},{y}_i,\mathbf{x}_i)$, minimizing $J'(\varrho^{(t+1)}, \bm{\pi},\tau)$ w.r.t. $\bm{\pi}$ leads to
\begin{align}\label{eqn:mean_new}
{\mu}_{\ell}^{(t+1)}=\frac{\sum_{i=1}^N\varrho(\ell|\bm{\pi}^{(t)},{y}_i,\mathbf{x}_i){y}_i}{\sum_{i=1}^N\varrho(\ell|\bm{\pi}^{(t)},{y}_i,\mathbf{x}_i)},
\end{align}
and
\begin{align} \label{eqn:covar_new}
{\sigma}_{\ell}^{(t+1)}=\frac{\sum_{i=1}^{N}\varrho(\ell|\bm{\pi}^{(t)},{y}_i,\mathbf{x}_i)({y}_i-{\mu}_{\ell}^{(t+1)})^2}{\sum_{i=1}^{N}\varrho(\ell|\bm{\pi}^{(t)},{y}_i,\mathbf{x}_i)}.
\end{align}
Comparing the two groups of update functions for leaf nodes, \emph{i.e.}, Eq.~\ref{eqn:mean} \emph{vs.} Eq.~\ref{eqn:mean_new} and Eq.~\ref{eqn:covar} \emph{vs.} Eq.~\ref{eqn:covar_new}, the only difference between them is that the posterior is changed from $\rho(\ell|{\bm{\pi}},{y}_i,\mathbf{x}_i)$ to $\varrho(\ell|{\bm{\pi}},{y}_i,\mathbf{x}_i)$. When $\tau \rightarrow 0$, the posterior $\varrho(\ell|{\bm{\pi}},{y}_i,\mathbf{x}_i)$ becomes uniform distribution, \emph{i.e.}, $\varrho(\ell|{\bm{\pi}},{y}_i,\mathbf{x}_i) \leftarrow \frac{1}{|\mathcal{L}|}$. Thus, for a small enough $\tau$, no matter how $\bm{\mu}^{(0)}$ and $\bm{\sigma}^{(0)}$ are initialized, $\varrho(\ell|{\bm{\pi}},{y}_i,\mathbf{x}_i)$ does not influence the optimization, since $J'(\varrho, \bm{\pi},\tau)$ always has the minimum: $\mu_{\ell}=\frac{1}{N}\sum_{i=1}^Ny_i$, $\sigma_{\ell}=\frac{1}{N}\sum_{i=1}^N(y_i-\mu_{\ell})^2$. Then, by gradually increasing $\tau$ (decreasing the temperature), $\varrho(\ell|{\bm{\pi}},{y}_i,\mathbf{x}_i)$ gradually becomes nonuniform, and the influence of $\varrho(\ell|{\bm{\pi}},{y}_i,\mathbf{x}_i)$ in the optimization is increasedly strengthened. In this annealing process, we also start with each input sample equally influencing all leaf nodes and gradually localize the influence, which is a well-known strategy to obtain better optima~\cite{Ref:Rose98deterministicannealing}.

\subsection{Learning a Regression Forest}
We use the same strategies in Sec.~\ref{sec:forest_learning_ldl} to learn a regression forest. A regression forest is an ensemble of regression trees $\mathcal {F}=\{\mathcal {T}^1,\ldots,\mathcal {T}^K\}$, where all trees can possibly share the same parameters in $\bm{\Theta}$, but each tree can have a different set of split functions (assigned by $\varphi$, as shown in Fig.~\ref{fig:DRForest}), and independent leaf node distribution $\bm{\pi}$. The loss function for a regression forest is also defined as the averaged loss functions of all individual trees (Eq.~\ref{eqn:forest_loss}). Learning the forest $\mathcal {F}$ also follows the alternating optimization strategy described in Sec.~\ref{sec:tree_opt_reg}. To learn $\bm{\Theta}$, we have the gradient of the forest loss given in Eq.~\ref{eqn:forest_loss}, but the first term of the right side is computed by Eq.~\ref{eqn:gradient_reg}. For the temperature parameter $T$ introduced when optimizing $\bm{\Theta}$, we use the same initialization and cooling schedule described in Sec.~\ref{sec:forest_learning_ldl}. The leaf node distribution $\bm{\pi}$ of each tree in the forest $\mathcal {F}$ is updated independently according to Eq.~\ref{eqn:mean_new} and Eq.~\ref{eqn:covar_new}. For the temperature parameter $\tau$ introduced in optimizing $\bm{\pi}$, we initialize it as a small value $\tau_0$ ($\tau_0 < 1$) and gradually increase it by $\tau \leftarrow \tau/\eta$ until $\tau=1$, where $\eta$ is a constant cooling factor less than 1. The training procedure of a DRF is shown in Algorithm.~\ref{fig:algorithm_reg}.
\begin{algorithm}
\caption{The training procedure of a DRF.}
\label{fig:algorithm_reg}
\begin{algorithmic}
\REQUIRE $\mathcal{S}$: training set
\REQUIRE $n_B$: the number of mini-batches to update $\bm{\pi}$
\REQUIRE $T_0, \tau_0$: two starting temperature parameters
\REQUIRE $\eta$: a constant cooling factor
\STATE Initialize $\bm{\Theta}$ and $\bm{\pi}$ randomly.
\STATE Set $\mathcal{B}=\{\emptyset\}$, $T=T_0$ and $\tau=\tau_0 $
\WHILE {Not converge}
\WHILE {$|\mathcal{B}|<n_B$}
\STATE Randomly select a mini-batch $B$ from $\mathcal{S}$
\STATE Update $\bm{\Theta}$ by Gradient Descent (Eq.~\ref{eqn:gradient_forest}, Eq.~\ref{eqn:gradient_reg}) on $B$
\STATE $\mathcal{B}=\mathcal{B}\bigcup{B}$
\ENDWHILE
\STATE Update $\bm{\pi}$ by iterating Eq.~\ref{eqn:mean_new} and Eq.~\ref{eqn:covar_new} on $\mathcal{B}$
\STATE $\mathcal{B}=\{\emptyset\}$
\IF {$T\geq 0$}
\STATE {$T \leftarrow \eta T$}
\ENDIF
\IF {$\tau < 1$}
\STATE {$\tau \leftarrow \tau/\eta$}
\ENDIF
\ENDWHILE
\end{algorithmic}
\end{algorithm}

In the testing stage, the output of the forest $\mathcal {F}$ is given by averaging the predictions from all the individual trees:
\begin{align}
&\hat{{y}} = \mathbf{g}(\mathbf{x};\mathcal {F})=\frac{1}{K}\sum_{k=1}^K\mathbf{g}(\mathbf{x};\mathcal {T}^k)\nonumber\\
&=\frac{1}{K}\sum_{k=1}^K\int{y}p({y}|\mathbf{x};\mathcal {T}^k)d{y}\nonumber\\
&=\frac{1}{K}\sum_{k=1}^K\int{y}\sum_{\ell\in\mathcal{L}^k}P(\ell|\mathbf{x};\bm{\Theta})\pi_{{\ell}}({y})d{y}\nonumber\\
&=\frac{1}{K}\sum_{k=1}^K\sum_{\ell\in\mathcal{L}^k}P(\ell|\mathbf{x};\bm{\Theta}){\mu}_{\ell},
\end{align}
where $\mathcal{L}^k$ is the leaf node set of the $k$-th tree. Here, we take the fact that the expectation of the Gaussian distribution $\pi_{\ell}({y})$ is ${\mu}_{\ell}$.

\section{Experiments}
In this section, we first introduce our experimental setup including the datasets, evaluation metrics, and implementation details. Then we compare our results
with others to show the effectiveness of our algorithms. After that, we conduct elaborate ablation experiments
to study the influence of different designs and variants of our methods.
Finally, we discuss the comparison between our DLDLF and
DRF, visualizations of the learned leaf nodes, the hyper-parameters and performance variance brought by random assignment $\varphi(\cdot)$.
\subsection{Experimental setup}
\subsubsection{Datasets}
\begin{figure}[!t]
\centering
\includegraphics[trim=3cm 5cm 4cm 5cm, clip=true, width=1.0\linewidth]{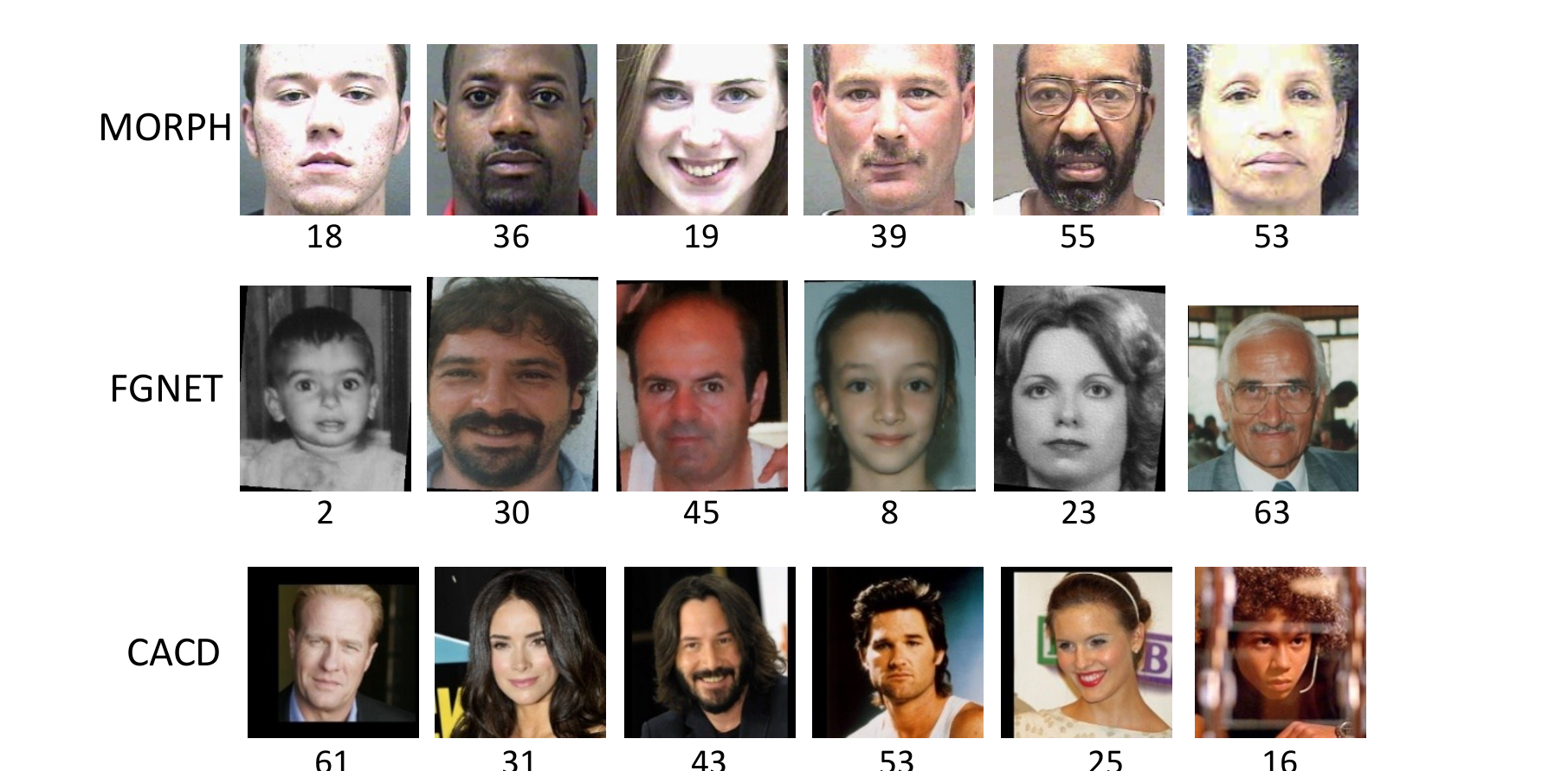}
\caption{Some examples of MORPH~\cite{Ref:MORPH06}, FG-NET~\cite{Ref:PanisLTC16} and CACD~\cite{Ref:ChenCH15}. The number below each image is the chronological age of each subject.}\label{fig:Example_Dataset}
\end{figure}

We conduct our experiments on three standard benchmarks: MORPH~\cite{Ref:MORPH06}, FG-NET~\cite{Ref:PanisLTC16} and the Cross-Age Celebrity Dataset (CACD)~\cite{Ref:ChenCH15}. Some examples of these three datasets are illustrated in Fig.~\ref{fig:Example_Dataset}.

\textbf{MORPH.}
MORPH is the most popular dataset for age estimation, which contains more than 55,000 images from about 13,000 people of different races. Each of the facial image is annotated with a chronological age.
The ethnicity of MORPH is very unbalanced, as more than 96\% of the facial images are from African or European people.

Existing methods adopted different experimental setups on MORPH. The first setup (Setup I)
is introduced in~\cite{Ref:Chang11,Ref:Chen2013Cumulative,Ref:Guo08,Ref:Wang2015Deeply,Ref:Rothe2016Some,Ref:Rothe16,Ref:Agustsson17}, which selects 5,492 images of Caucasian people
from the original MORPH dataset, to reduce the cross-ethnicity effects.
In Setup I, these 5,492 images are randomly partitioned into two subsets:
80\% of the images are selected for training and others for testing.
The random partition is repeated 5 times, and the final performance is averaged over these 5 different partitions.

The second setup (Setup II) is used in~\cite{Ref:Geng2010Facial,Ref:geng2016label,Ref:Gao2017Deep}, under which all of the images in MORPH are randomly split into training/testing ($80\%$/$20\%$) sets.
The random splitting is performed 5 times repeatedly and the final performance is obtained by averaging the performances of these 5 different splits.

There are also several methods~\cite{Ref:GuoM11,Ref:guo2014framework,Ref:yi2014age} using the third setup (Setup III), which randomly selected a subset (about 21,000 images) from MORPH and restricted the ratio between Black and White and the one between Female and Male are 1:1 and 1:3, respectively.

\textbf{FG-NET.}
FG-NET~\cite{Ref:PanisLTC16} is also a widely used dataset for age estimation. It contains 1002 facial images of 82 individuals, in which most of them are white people. Each individual in FG-NET has more than 10 photos taken at different ages. The images in FG-NET have a large variation in lighting conditions, poses and expressions.

Following the experimental setup used in~\cite{Ref:Yan2007Learning,Ref:Guo08,Ref:ChangCH11,Ref:Chen2013Cumulative,Ref:Rothe16}, we perform ``leave one out'' cross validation on this dataset, \emph{i.e.}, we leave images of one person for testing and take the remaining images for training.

\textbf{CACD.}
CACD~\cite{Ref:ChenCH15} is a large dataset which has around 160,000 facial images of 2,000 celebrities collected from the Internet. These celebrities are divided into three subsets: the training set, the testing set and the validation set which consist of 1,800 celebrities, 120 celebrities and 80 celebrities, respectively. The validation and testing sets are clean but the training set is noisy.

For evaluation we adopt the setup used in~\cite{Ref:Rothe16}. They report results on the testing set obtained by using the models trained on the training set and the validation set, respectively.

\subsubsection{Evaluation metric}
The performance of age estimation is evaluated in terms of mean absolute error (MAE) as well
as Cumulative Score (CS).
MAE is the average absolute error over the testing set,
and the Cumulative Score is calculated by $\text{CS}(l) = \frac{K_l}{K} \cdot 100\%$, where $K$ is the total number of testing images and $K_l$ is the number of testing facial images whose absolute error between the estimated age and the ground truth age is not greater than $l$ years. Here, we set the same error level 5 as in~\cite{Ref:Chang11,Ref:Chen2013Cumulative,Ref:huerta2014facial}, \emph{i.e.}, $l=5$. Note that, since not all the methods reported the Cumulative Score, we are only able to give CS values for some competitors.

\subsubsection{Implementation details}
Our realizations of DLDLFs and DRFs are based on the public available ``caffe''~\cite{Ref:Jia2014caffe} framework. Following recent deep learning based age estimation methods~\cite{Ref:Rothe16,Ref:Rothe2016Some,Ref:Agustsson17,Ref:Gao2017Deep}, we use the VGG-16 Net~\cite{Ref:Simonyan14} as the CNN part of the proposed DLDLFs and DRFs.

\textbf{Parameters Setting.} The forest model related hyper-parameters (and the default values we used)
are:
number of trees (5), tree depth (6),
number of output units produced by the feature learning function (128),
iterations to update leaf-node predictions (20),
number of mini-batches used to update leaf node predictions (50).

The age distribution generation related hyper-parameter (and the default value we used) is: the pre-defined standard deviation $\alpha$ in Eq.~\ref{eqn:age_distribution} (2.0).

The network training related hyper-parameters (and the values we used) are: initial learning rate (0.05), mini-batch size (16), maximal iterations (30k). We decrease the learning rate ($\times 0.5$) every 10k iterations.

We fixed the values for the temperature parameters introduced in our DA processes: $T_0=1$, $\tau_0=0.5$ and $\eta = 0.9$.

\textbf{Preprocessing.}
Face alignment is a common preprocessing operation for age estimation~\cite{Ref:yi2014age, Ref:Rothe16,Ref:Niu16,Ref:Agustsson17,Ref:ChenZDLR17,Ref:Gao2017Deep}.

Following these previous methods, we perform face alignment to guarantee all eyeballs stay at the same position in the image: Faces are firstly detected by
using a standard face detectior~\cite{Ref:ViolaJ01} and facial landmarks are localized by AAM~\cite{Ref:CootesET98}.

\subsection{Performance Comparison}\label{sec:performance_comp}
In this section we compare our LDLF and DRF with other state-of-the-art age estimation methods on the three standard benchmarks: MORPH~\cite{Ref:MORPH06}, FG-NET~\cite{Ref:PanisLTC16} and the Cross-Age Celebrity Dataset (CACD)~\cite{Ref:ChenCH15}.

\textbf{MORPH.}
We first compare the proposed LDLF and DRF with other state-of-the-art age estimation methods on MORPH. As we described before, there are three experimental setups used on this dataset. For a fair comparison, we test the proposed LDLF and DRF on MORPH under all these three setups. The quantitative results of the three settings are summarized in Table~\ref{table:morph-setup1}, Table~\ref{table:morph-setup2} and Table~\ref{table:morph-setup3}, respectively. As can be seen from these tables, our DRF and LDLF achieve the best and the second best performance on all of the setups, respectively, and outperform the current state-of-the-arts with a clear margin. This result shows the effectiveness of jointly learning input-dependent data partition and data distributions in local partitions for age estimation.


\begin{table}
\begin{center}
\begin{tabular}{l|c|c}
\hline
Method & MAE & CS  \\
\hline\hline
Human workers ~\cite{Ref:HanOLJ15} & 6.30 & 51.0 \%*\\
AGES ~\cite{Ref:geng2007automatic} & 8.83& 46.8 \%*\\
MTWGP ~\cite{Ref:zhang2010multi} & 6.28 & 52.1\%*\\
CA-SVR ~\cite{Ref:Chen2013Cumulative} & 5.88& 57.9\% \\
SVR ~\cite{Ref:Guo08} & 5.77 & 57.1\%\\
DLA ~\cite{Ref:Wang2015Deeply} & 4.77 & 63.4 \%*\\
Rank ~\cite{Ref:chang2010ranking} & 6.49& 49.5\%\\
Rothe ~\cite{Ref:Rothe2016Some} & 3.45&  N/A\\
DEX ~\cite{Ref:Rothe16} & 3.25 &N/A\\
ARN ~\cite{Ref:Agustsson17} & 3.00&N/A\\
\hline
DLDLF (ours) & 2.94& 84.7\%\\
DRF (ours) &\textbf{2.80} & \textbf{85.6\%}\\
\hline
\end{tabular}
\end{center}
\caption{Performance comparison on MORPH~\cite{Ref:MORPH06} (Setup I)(*: the value is read from the CS curve shown in the reference).}
\label{table:morph-setup1}
\end{table}

\begin{table}
\begin{center}
\begin{tabular}{l|c|c}
\hline
Method & MAE& CS  \\
\hline\hline
IIS-LDL ~\cite{Ref:Geng2010Facial} & 5.67&71.2\%*\\
CPNN ~\cite{Ref:Geng13} & 4.87& N/A\\
Huerta ~\cite{Ref:huerta2014facial} & 4.25 &71.2\%\\
BFGS-LDL ~\cite{Ref:geng2016label} & 3.94& N/A\\
OHRank ~\cite{Ref:Chang11} & 6.07&56.3\%\\
OR-SVM ~\cite{Ref:chang2010ranking} & 4.21&68.1\%*\\
CCA ~\cite{Ref:guo2013joint}  & 4.73&60.5\%*\\
LSVR ~\cite{Ref:GuoMFH09}  & 4.31&66.2\%*\\
OR-CNN ~\cite{Ref:Niu16} & 3.27& 81.5\%\\
SMMR ~\cite{Ref:Huang17} & 3.24& N/A\\
Ranking-CNN ~\cite{Ref:ChenZDLR17} & 2.96 & 85.2\%\\
DLDL ~\cite{Ref:Gao2017Deep} & 2.42& N/A\\
Mean-Variance Loss ~\cite{Ref:Pan18} & 2.41 & 91.2\% \\
\hline
DLDLF (ours) & 2.19& \textbf{93.0\%}\\
DRF (ours) &\textbf{2.14}&91.3\%\\
\hline
\end{tabular}
\end{center}
\caption{Performance comparison on MORPH~\cite{Ref:MORPH06} (Setup II)(*: the value is read from the CS curve shown in the reference).}
\label{table:morph-setup2}
\end{table}

\begin{table}
\begin{center}
\begin{tabular}{l|l|l}
\hline
Method & MAE & CS\\
\hline\hline
KPLS ~\cite{Ref:GuoM11} & 4.18 & N/A \\
Guo and Mu ~\cite{Ref:guo2014framework} & 3.92 & N/A\\
CPLF~\cite{Ref:yi2014age} & 3.63 & N/A\\
\hline
DLDLF (ours) & 2.99& \textbf{85.6\%}\\
DRF (ours) &\textbf{2.90}& 82.7\%\\
\hline
\end{tabular}
\end{center}
\caption{Performance comparison on MORPH~\cite{Ref:MORPH06} (Setting III).}
\label{table:morph-setup3}
\end{table}

\textbf{FG-NET.}
We then conduct experiments on FG-NET~\cite{Ref:PanisLTC16}. The quantitative comparisons on FG-NET dataset are shown in Table~\ref{table:fg-net}. As can be seen, DRF and DLDLF outperform other methods significantly. Note that, they are the only two methods that have a MAE below 4.0. The age distribution of FG-NET is strongly biased, moreover, the ``leave one out'' cross validation policy further aggravates the bias between the training set and the testing set.
The ability of overcoming the bias between training and testing sets indicates that the proposed
LDLF and DRF can handle inhomogeneous data well.

\begin{table}[htb]
\begin{center}
\begin{tabular}{l|c|c}
\hline
Method & MAE & CS \\
\hline\hline
Human workers ~\cite{Ref:HanOLJ15} & 4.70 & 69.5\%* \\
Rank ~\cite{Ref:chang2010ranking} & 5.79& 66.5\%*\\
DIF ~\cite{Ref:HanOLJ15}  & 4.80 &74.3\%*\\
AGES ~\cite{Ref:geng2007automatic} & 6.77& 64.1\%*\\
IIS-LDL ~\cite{Ref:Geng2010Facial} & 5.77&N/A\\
CPNN ~\cite{Ref:Geng13} & 4.76&N/A\\
MTWGP ~\cite{Ref:zhang2010multi}  & 4.83& 72.3\%*\\
CA-SVR ~\cite{Ref:Chen2013Cumulative} & 4.67& 74.5\%\\
LARR ~\cite{Ref:Guo08}  & 5.07 & 68.9\%*\\
OHRank ~\cite{Ref:Chang11} & 4.48& 74.4\%\\
DLA ~\cite{Ref:Wang2015Deeply}  & 4.26 &N/A\\
CAM~\cite{Ref:luu2011contourlet} & 4.12 & 73.5\%*\\
Rothe ~\cite{Ref:Rothe2016Some}  & 5.01& N/A\\
DEX ~\cite{Ref:Rothe16}  & 4.63&N/A\\
Mean-Variance Loss ~\cite{Ref:Pan18} & 4.10 & 78.5\%* \\
\hline
DLDLF (Ours) &3.71 & 84,8\%\\
DRF (Ours) &\textbf{3.47} & \textbf{87.3\%}\\
\hline
\end{tabular}
\end{center}
\caption{Performance comparison on FG-NET~\cite{Ref:PanisLTC16}(*: the value is read from the CS curve shown in the reference).}
\label{table:fg-net}
\end{table}

\textbf{CACD.}
Finally, we conduct our experiments on CACD~\cite{Ref:ChenCH15}. The detailed comparisons are shown in Table~.\ref{table:cacd}. The proposed DLDLF and DRF perform better than the competitor DEX~\cite{Ref:Rothe16}, no matter which set they are trained on. It's worth noting that, the improvements of DLDLF and DRF to DEX are much more significant when they are trained on the validation set than the training set. This result can be explained as follow: As described earlier, the inhomogeneous data is the main challenge for training age estimation models. This challenge can be alleviated by enlarging the number of the training samples. Therefore, DEX, DLDLF and DRF achieve comparable results when they are trained on the training set. But when they are trained on the validation set, which is much smaller than the training set, DLDLF and DRF, especially DRF, outperform DEX significantly, because DLDLF and DRF directly address the inhomogeneity challenge. Therefore, DLDLF and DRF are capable of handling inhomogeneous data even when learned from a small set.

\begin{table}
\begin{center}
\begin{tabular}{l|c|c|c}
\hline
Trained on & Dex~\cite{Ref:Rothe16} & DLDLF (Ours)& DRF (Ours) \\
\hline\hline
CACD (train)  & 4.785 & 4.679 &\textbf{4.610}\\
CACD (val) &  6.521& 6.162&\textbf{5.630}\\
\hline
\end{tabular}
\end{center}
\caption{Performance comparison on CACD (measured by MAE)~\cite{Ref:ChenCH15}.}
\label{table:cacd}
\end{table}

\subsection{Ablation Study}
We conduct some ablation experiments on the MORPH dataset (Setup I), to analyze the influence of different designs and components for our methods. We want to answer these questions from the ablation study: (1) Since we argue that age estimation is usually formulated as an LDL or regression problem rather than a classification problem, what the result would be if we addressed age estimation by a deep classification forest model~\cite{Ref:Kontschieder15}? (2) Since we argue that our forest structure is important for age estimation, what the result would be if we replaced our forest structure by an $\ell_2$ regression loss function? (3) Since we argue that the convergence of the update functions for leaf nodes used in NRF~\cite{Ref:Roy16} is not guaranteed, what the result would be if we changed our update functions in DRFs to them? (4) What the result would be if the DA process for learning split nodes was not used ($T=0$)? (5) What would the result be if the DA process for learning leaf nodes in DRFs was not used ($\tau=1$), especially when leaf nodes are not initialized by kmeans clustering (w/o kmeans initialization)? To answer these questions, we consider these variants of our methods in the ablation experiments:
\begin{itemize}
\item DLDLF ($T=0$): the DLDLF \textbf{without} the DA process for learning split nodes, \emph{i.e.}, the baseline DLDLFs proposed in our previous work~\cite{Ref:Shen17}.
\item DLDLF (full model): the DLDLF \textbf{with} the DA process for learning split nodes, \emph{i.e.}, the method described in Sec.~\ref{sec:DLDLFs}.
\item DRF ($T=0$, $\tau=1$, w/ kmeans initialization): the DRF \textbf{without} the two DA processes for learning split and leaf nodes and \textbf{with} kmeans initialization for leaf nodes, \emph{i.e.}, the baseline DRF proposed in our previous work~\cite{Ref:Shen18}.
\item DRF ($T=0$, $\tau=1$): the DRF \textbf{without} the two DA processes for learning split and leaf nodes and also \textbf{without} kmeans initialization for leaf nodes.
\item DRF ($\tau=1$, w/ kmeans initialization): the DRF \textbf{with} the DA process for learning split nodes, \textbf{without} the two DA processes for leaf nodes, and \textbf{with} kmeans initialization for leaf nodes.
\item DRF ($T=0$): the DRF \textbf{without} the DA process for learning split nodes, \textbf{with} the two DA processes for leaf nodes, and \textbf{without} kmeans initialization for leaf nodes.
\item DRF (full model): the DRF \textbf{with} the two DA processes for learning split and leaf nodes and \textbf{without} kmeans initialization for leaf nodes, \emph{i.e.}, the method described in Sec.~\ref{sec:DRFs}.
\end{itemize}
The results of the ablation experiments are summarized in Table~\ref{table:ablation-study-setup1}.
\subsubsection{Age estimation by classification}
To formulate age estimation as a classification problem, we treat each age value as a class. We apply a deep classification forest model, deep neural decision forests (DNDFs)~\cite{Ref:Kontschieder15}, to address this problem. As shown in Table~\ref{table:ablation-study-setup1}, the age estimation result obtained by DNDFs is much worse than the results obtained by our baseline DLDLF and DRF, \emph{i.e.}, DLDLF ($T=0$) and DRF ($T=0,\tau=1$, w/ kmeans initialization), which evidences that age estimation is not suitable to be formulated as a classification problem.
\subsubsection{Age estimation w/o forest structure}
To verify whether our forest structure is important for age estimation, we replace our forest structure by an $\ell_2$ norm (Euclidean) loss function and denote this method by Deep Regression. As shown in Table~\ref{table:ablation-study-setup1}, the age estimation result obtained by Deep Regression is even worse than the result obtained by our baseline DRF, \emph{i.e.}, DRF ($T=0,\tau=1$, w/ kmeans initialization), which evidences the importance of our forest structure.
\subsubsection{The update functions for leaf nodes}
The method which replaces the update functions for leaf nodes in a NRF by those in a DRF is denoted by DRF-NRF. For fair comparison, we also initialize leaf node distributions in DRF-NRF by kmeans.
As can be seen, using NRF's leaf node update functions leads to a much worse result than our baseline DRF, \emph{i.e.}, DRF ($T=0,\tau=1$, w/ kmeans initialization), which agrees with our concern about the convergence of NRF's leaf node update rule.
\subsubsection{The DA process for learning split nodes}
\begin{figure}[!t]
\centering
\includegraphics[trim=0.5cm 0cm 1.2cm 0cm, clip=true, width=1.0\linewidth]{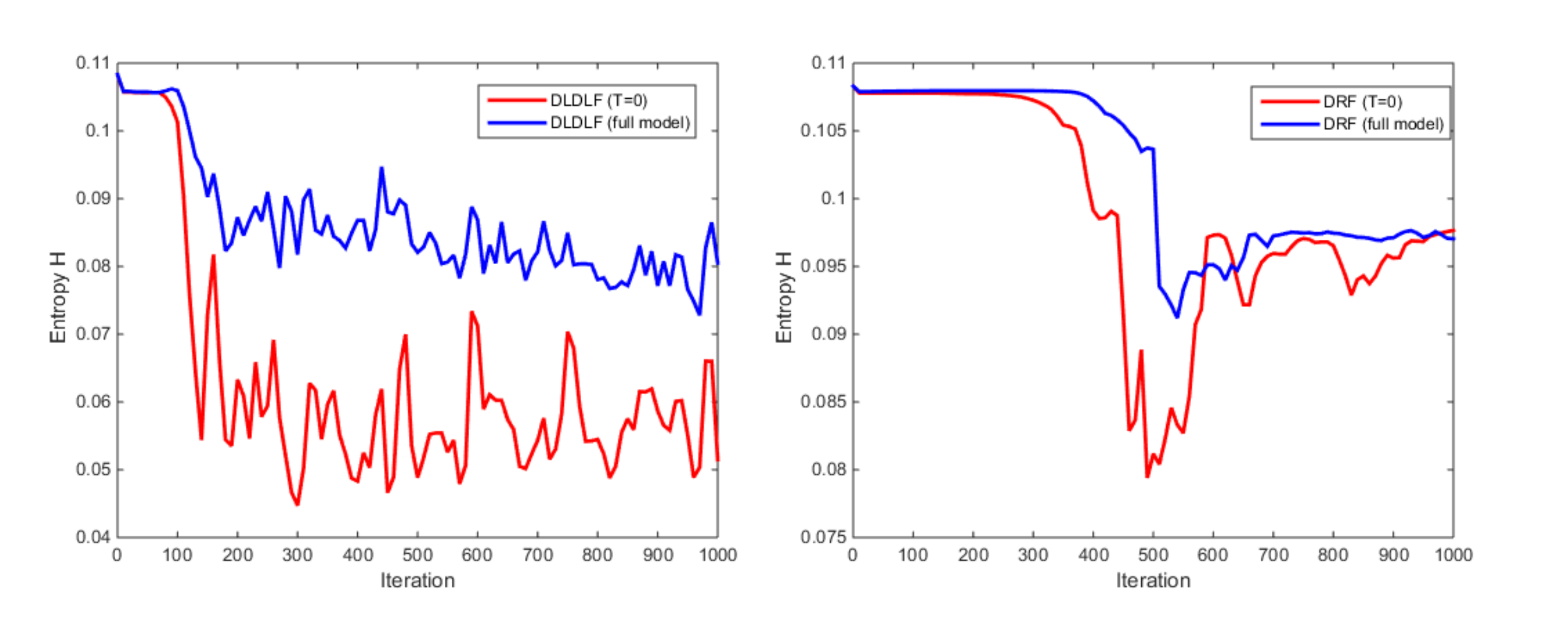}
\caption{The dynamics of the averaged entropy $H(\bm{\Theta}; \mathcal{S})$ over the training
set $\mathcal{S}$ for DLDLFs (left) and DRFs (right) at the beginning of tree learning.}\label{fig:dynamics_DA}
\end{figure}

By comparing DLDLF (full model), DRF (full model) and DRF ($\tau=1$, w/ kmeans initialization) with DLDLF ($T=0$), DRF ($T=0$) and DRF ($T=0$, $\tau=1$, w/ kmeans initialization), respectively, we see that using the DA process for learning split nodes always leads to better performances. We also show the dynamics of the averaged entropy $H(\bm{\Theta};\mathcal{S})$ over the training set $\mathcal{S}$ at the beginning of tree learning
in Fig.~\ref{fig:dynamics_DA}, where we see for both DLDLF (full model) and
DRF (full model), the averaged entropies $H(\bm{\Theta};\mathcal{S})$ of them
are lager and decrease more slowly than those of DLDLF
($T = 0$) and DRF ($T = 0$). This result is consistent with
our intuition about the DA process for learning split nodes
described in Sec.~\ref{sec:split_node_ldl}.

\subsubsection{The DA process for learning leaf nodes}
We first compare DRF ($T=0$, $\tau=1$, w/ kmeans initialization) with DRF ($T=0$, $\tau=1$, w/o kmeans initialization). The result obtained by DRF ($T=0$, $\tau=1$, w/o kmeans initialization) is much worse than the one obtained by DRF ($T=0$, $\tau=1$, w/ kmeans initialization). This comparison shows that without the DA process for learning leaf nodes, the leaf node update may converge to a poor local minimum, if the leaf node parameters are not well initialized. We then compare DRF (full model) with DRF ($\tau=1$, w/ kmeans initialization), where we see that using the DA process for learning leaf nodes, even without well initializing the leaf node parameters, can lead to a better performance. This comparison indicates that the proposed DA process for learning leaf nodes can avoid poor local optima and obtain better estimates of tree parameters free of initial parameter values.
\begin{table}[!h]
\footnotesize
\begin{center}
\begin{tabular}{l|c|c}
\hline
Method & MAE  & CS\\
\hline\hline
DNDF~\cite{Ref:Kontschieder15} & 3.32 & 80.7\% \\
Deep Regression & 3.21 & 81.1\%\\
DRF-NRF (w/ kmeans initialization)~\cite{Ref:Roy16} & 3.51& 75.4\%\\
\hline
DLDLF ($T=0$)   & 3.02 &81.3\%\\
DLDLF (full model) & 2.94 & 84.7\%\\
DRF ($T=0$, $\tau=1$, w/ kmeans initialization)  & 2.91& 82.9\%\\
DRF ($T=0$, $\tau=1$)  & 6.91 & 47.8\%\\
DRF ($\tau=1$, w/ kmeans initialization)  & 2.85 & 83.9\%\\
DRF ($T=0$) & 2.85 & 84.1\%\\
DRF (full model)  & 2.80 & 85.6\%\\
\hline
\end{tabular}
\end{center}
\caption{Ablation study on MORPH [43] (Setup I).}
\label{table:ablation-study-setup1}
\end{table}

\subsection{Discussion}
\subsubsection{Comparison between DLDLFs and DRFs}
Although DLDLFs and DRFs formulate age estimation as
different problems, both of them take the strong correlation
between close ages of the same individual into account. The
difference is DLDLFs explicitly model cross age correlations
of the same individual, while DRFs do this implicitly. The
experimental results in Sec.~\ref{sec:performance_comp} show that DRFs always
achieve lower (better) MAE than DLDLFs. The reason might
be that DRFs directly approach the precise chronological
ages. However, an interesting result is, for both Setup II
and Setup III of the MORPH dataset, that DLDLFs obtain
a worse MAE than DRFs, but a better CS. This is because
the CS metric tolerates small prediction errors, and DLDLFs
learn an age distribution, which benefits fuzzy prediction.
Another reason for the worse MAEs achieved by DLDLFs
is that we generate age distributions by a fixed $\alpha$ for
different individuals, which might be inflexible in adapting
to complex face data domains with diverse cross-age correlations~\cite{Ref:He17}. Our future work is to learn the age distributions
adaptive to different individuals~\cite{Ref:Pan18}.
\subsubsection{Visualization of learned leaf nodes}
\begin{figure*}[!t]
\centering
\includegraphics[trim=0cm 0cm 0cm 0.6cm, clip=true, width=1.0\linewidth]{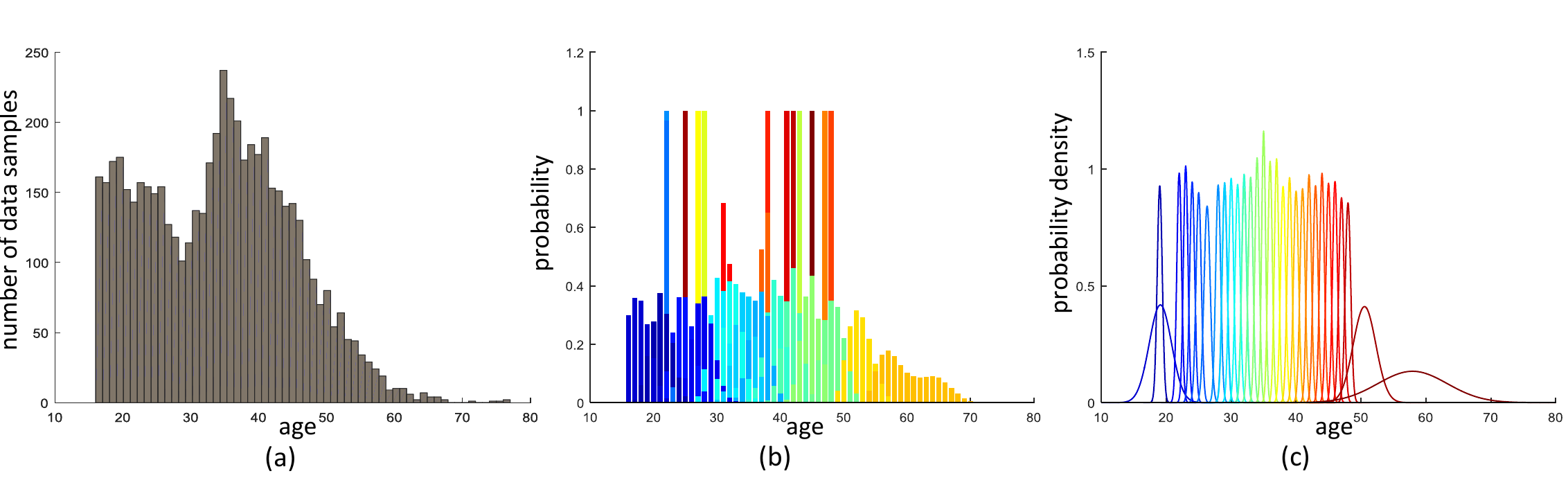}
\caption{(a) Histogram of data samples with respect to age on MORPH~\cite{Ref:MORPH06} (Setup I). (b) Visualization of the learned leaf node distributions in our DLDLF. (c) Visualization of the learned leaf node distributions in our DRF. The distributions held by different leaf nodes are in different (gradually varied) colors, which are best viewed in color.}\label{fig:LeafNodeDistr}
\end{figure*}

To better understand DLDLF and DRF, we visualize the distributions at the leaf nodes learned on MORPH~\cite{Ref:MORPH06} (Setup I) in Fig.~\ref{fig:LeafNodeDistr}(b) and Fig.~\ref{fig:LeafNodeDistr}(c), respectively. For reference, we also display the histogram of data samples (the vertical axis) with respect to age (the horizontal axis). Each leaf node in our DLDLF contains a discrete probability distribution, which is represented by a colored histogram in Fig.~\ref{fig:LeafNodeDistr}(b). Each leaf node in our DRF contains a Gaussian distribution, as visualized in in Fig.~\ref{fig:LeafNodeDistr}(c). The horizontal axes in both Fig.~\ref{fig:LeafNodeDistr}(b) and Fig.~\ref{fig:LeafNodeDistr}(c) represent age and the vertical axes in them represent probability and probability density, respectively (we rescale some very sharp distributions in Fig.~\ref{fig:LeafNodeDistr}(c) for better visualization, since the peak densities of them are too large). According to Fig.~\ref{fig:LeafNodeDistr}(a), the age data in MORPH was sampled mostly below age 60, and densely concentrated around 20's and 40's. As shown in Fig.~\ref{fig:LeafNodeDistr}(b) and Fig.~\ref{fig:LeafNodeDistr}(c), both of the distributions learned by DLDLF and DRF fit the age data well: The discrete distributions around 60 are more uniform than those spread in the interval between 20 and 50; The Gaussian distribution centered around 60 has much larger variance than those centered in the interval between 20 and 50, but has smaller probability density. Note that, although these learned distributions represent homogeneous local partitions, the number of samples is not necessarily uniformly distributed among partitions. Another phenomenon is these distributions are heavily overlapped, which accords with the fact that different people with the same age but have quite different facial appearances.

\subsubsection{Sensitivity of hyper-parameters} \label{sec:param_discussion}
Now we discuss three important hyper-parameters: the tree number, the tree depth and
the standard deviation $\alpha$ in Eq.~\ref{eqn:age_distribution} used for age distribution generation. We vary each of them and fix the other one to the default value to see how the performance changes on MORPH (Setup I).

\begin{figure*}[!ht]
\centering
\includegraphics[trim=0cm 0cm 0cm 0cm, clip=true, width=1.0\linewidth]{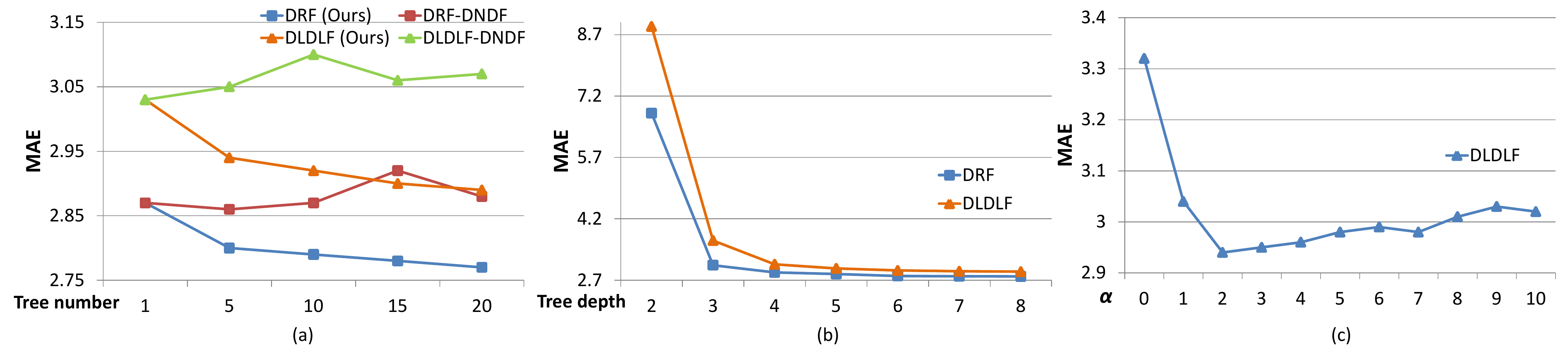}
\caption{Performance changes by varying (a) tree number, (b) tree depth and (c) standard deviation $\alpha$ on MORPH~\cite{Ref:MORPH06} (Setup I).}\label{fig:ParameterDiscussion}
\end{figure*}

\textbf{Tree number.}
As we stated in Sec.~\ref{sec:intro}, the ensemble strategy to learn a forest proposed in DNDFs~\cite{Ref:Kontschieder15} is different from ours. Therefore, it is necessary to see which ensemble strategy is better to learn a forest. Towards this end, we replace our ensemble strategy by the one used in DNDFs, and name the methods DLDLF-DNDF and DRF-DNDF, accordingly. As shown in Fig.~\ref{fig:ParameterDiscussion}(a), our
ensemble strategy can improve the performance by using more trees, as we expected, while the one used in DNDFs leads to an even worse performance than one for a single tree.

\textbf{Tree depth.}
Tree depth is another important parameter for decision trees. As shown in Fig.~\ref{fig:ParameterDiscussion} (b), for both DLDLF and DRF, with the tree depth increase, the MAE first becomes lower and then stable. One concern about the tree depth is that a very deep tree may lead to underflow due to the continued product form of Eq.~\ref{eqn:prob_leaf}. However, this will not happen in practice. First, according to Fig.~\ref{fig:ParameterDiscussion} (b), the performance of our method becomes saturated when the tree depth is larger than 6, thus it is unnecessary to use very deep trees. Second, there is an implicit constraint between tree depth $h$ and unit number $m$ of the FC layer $\mathbf{f}$: $m \geq 2^{h-1} - 1$. The maximum unit number of the FC layers in a typical CNN architecture, \emph{e.g.}, AlexNet~\cite{Ref:KrizhevskySH12} and VGG-16 Net~\cite{Ref:Simonyan14}, is 4096, which implies that the tree depth should not be larger than 13.

\textbf{Standard deviation $\alpha$.}
Fig.~\ref{fig:ParameterDiscussion} (c) shows that how the MAE of our DLDLF changes by using Gaussian distributions with different standard deviation $\alpha$ to generate age distributions. Note that, when $\alpha=0$, the generated age distributions are one-hot, \emph{i.e.}, the LDL problem becomes a standard classification problem, which leads to a significant performance reduction. When $\alpha$ is larger, the generated age distributions are dispersive, which also leads to performance decrease. This is consistent with our intuition that neighboring ages can help to describe the face appearance of a given age  but should not change the priority of the original given age.
\subsubsection{Performance variance brought by random assignment $\varphi(\cdot)$}
In our forests, CNN features are randomly selected and assigned to split nodes for forest training, as defined by the function $\varphi(\cdot)$. In order to to check whether this randomness will seriously affect performance, we train 50 DRFs on MORPH (Setup I) and find that the standard derivation of the results obtained by the 50 DRFs is only 0.01 MAE. Therefore, the random feature selection and assignment process does not seriously affect the performance. This result is plausible, since CNN features are first initialized by the values computed by random weights, or weights from a pre-trained model trained for classification on Imagenet, which is a very different task than age estimation, and then the selected CNN features used in the forest will be learned and optimized for age estimation during forest training.

\section{Conclusion}
We proposed two Deep Differentiable Random Forests, \emph{i.e.}, Deep Label Distribution Forest (DLDLF) and Deep Regression Forest (DRF) for age estimation, which learn nonlinear mapping between inhomogeneous facial feature space and ages. In these two forests, by performing soft data partition at split nodes, the forests can be connected to a deep network and learned in an end-to-end manner, where data partition at split nodes is learned by Back-propagation and age distribution at leaf nodes is optimized by iterating a step-size free and fast-converged update rule derived from Variational Bounding. In addition, two Deterministic Annealing processes are introduced into the learning of split and leaf nodes, respectively, to avoid poor local optima and obtain better estimates of tree parameters free of initial parameter values. The end-to-end learning of split and leaf nodes ensures that partition function at each split node is input-dependent and the local input-output correlation at each leaf node is homogeneous. Experimental results showed that DLDLF and DRF achieved state-of-the-art results on three age estimation benchmarks. Our Deep Differentiable Random Forests are also applicable to other problems with inhomogeneous
data, which will be investigated in our future work.

\appendix{}
In the appendix, we give a re-derivation of tree optimization
in DNDFs~\cite{Ref:Kontschieder15} from our perspective, \emph{i.e.,} to reproduce the
update functions for leaf nodes in DNDFs by Variational
Bounding. For a sample $\mathbf{x} \in \mathcal{X}$, its class label is $y \in \mathcal{Y}$, then
the output of the tree is given by averaging leaf predictions
by the probability of reaching the leaf:
\begin{align}
p({y}|\mathbf{x};\mathcal {T})=\sum_{\ell\in\mathcal{L}}P(\ell|\mathbf{x};\bm{\Theta})\pi_{{\ell}_y}.
\end{align}
where $\pi_{{\ell}_y}$ is the probability for class $y$ assigned by $\bm{\pi}_{\ell}$. Given
a training set $\mathcal {S}=\{(\mathbf{x}_i,y_i)\}_{i=1}^N$, the classification loss is
defined by
\begin{align}\label{eqn:tree_loss_cls}
R(\bm{\pi},\bm{\Theta};\mathcal{S})=-\frac{1}{N}\sum_{i=1}^{N}\log(p({y}_{i}|\mathbf{x}_{i}, \mathcal {T}))\nonumber\\
=-\frac{1}{N}\sum_{i=1}^{N}\log\big(\sum_{\ell\in\mathcal{L}}P(\ell|\mathbf{x}_i;\bm{\Theta})\pi_{{\ell}_{{y}_i}}\big).
\end{align}

By fixing $\bm{\Theta}$, we can obtain a tight upper bound of
$R(\bm{\pi}, \bm{\Theta}; \mathcal{S})$ by Jensen's inequality:

\begin{align}\label{eqn:jensen_cls}
R(\bm{\pi},\bm{\Theta};\mathcal{S})=-\frac{1}{N}\sum_{i=1}^{N}\log\big(\sum_{\ell\in\mathcal{L}}P(\ell|\mathbf{x}_i;\bm{\Theta})\pi_{{\ell}_{{y}_i}}\big)\nonumber\\
\leq  -\frac{1}{N}\sum_{i=1}^{N}\sum_{\ell\in\mathcal{L}}\rho(\ell|\bar{\bm{\pi}},\mathbf{x}_i)\log\Big(\frac{P(\ell|\mathbf{x}_i;\bm{\Theta})\pi_{{\ell}_{{y}_i}}}{\rho(\ell|\bar{\bm{\pi}},\mathbf{x}_i)}\Big)\nonumber\\
\end{align}
where $\rho(\ell|{\bm{\pi}},\mathbf{x}) = \frac{P(\ell|\mathbf{x};\bm{\Theta})\pi_{{\ell}_{{y}}}}{ \sum_{\ell\in\mathcal{L}}P(\ell|\mathbf{x}_i;\bm{\Theta})\pi_{{\ell}_{{y}}} }$ and when $\bm{\pi}=\bar{\bm{\pi}}$ the equality holds. This tight upper bound for $R(\bm{\pi}, \bm{\Theta}; \mathcal{S})$
indicates the optimization of parameter $\bm{\pi}$ by Variational
Bounding~\cite{Ref:Jordan99,Ref:Yuille03}. Let us rewrite the upper bound in Eq.~\ref{eqn:jensen_cls} as
\begin{align}\label{eqn:phi_cls}
\phi(\bm{\pi},\bar{\bm{\pi}}) = -\frac{1}{N}\sum_{i=1}^{N}\sum_{\ell\in\mathcal{L}}\rho(\ell|\bar{\bm{\pi}},\mathbf{x}_i)\log\Big(\frac{P(\ell|\mathbf{x}_i;\bm{\Theta})\pi_{{\ell}_{{y}_i}}}{\rho(\ell|\bar{\bm{\pi}},\mathbf{x}_i)}\Big).
\end{align}
Note that, $\phi(\bm{\pi},\bar{\bm{\pi}})$ has the properties that for any $\bm{\pi}$ and $\bar{\bm{\pi}}$, $\phi(\bm{\pi},\bar{\bm{\pi}})\geq \phi(\bm{\pi},\bm{\pi})=R(\bm{\pi},\bm{\Theta};\mathcal {S})$ and $\phi(\bar{\bm{\pi}},\bar{\bm{\pi}})= R(\bar{\bm{\pi}},\bm{\Theta};\mathcal {S})$. These two properties satisfy the conditions
for Variational Bounding. According to Variational Bounding,
an original objective function (\emph{e.g.}, $R(\bm{\pi}, \bm{\Theta}; \mathcal{S})$) to be
minimized gets replaced by its bound (\emph{e.g.}, $\phi(\bm{\pi},\bar{\bm{\pi}})$) in
an iterative manner. Assume that we are at a point $\bm{\pi}^{(t)}$ corresponding to the $t$-th iteration, then $\phi(\bm{\pi},\bm{\pi}^{(t)})$ is a tight upper bound for $R(\bm{\pi},\bm{\Theta};\mathcal {S})$. In the next iteration, $\bm{\pi}^{(t+1)}$ is chosen such that $\phi(\bm{\pi}^{(t+1)},\bm{\pi})\leq R(\bm{\pi}^{(t)},\bm{\Theta};\mathcal {S})$, which implies $R(\bm{\pi}^{(t+1)},\bm{\Theta};\mathcal {S})\leq R(\bm{\pi}^{(t)},\bm{\Theta};\mathcal {S})$. Consequently, we can minimize  $\phi(\bm{\pi},\bar{\bm{\pi}})$ instead of $R(\bm{\pi},\bm{\Theta};\mathcal {S})$ after ensuring that
$R(\bm{\pi}^{(t)},\bm{\Theta};\mathcal {S})=\phi(\bm{\pi}^{(t)},\bar{\bm{\pi}})$, \emph{i.e.}, $\bar{\bm{\pi}} = \bm{\pi}^{(t)}$. Thus, we
have
\begin{align}
\bm{\pi}^{(t+1)}=\arg\min_{\bm{\pi}}\phi(\bm{\pi},\bm{\pi}^{(t)}), \textbf{s.t.}, \forall\ell,\sum_y\pi_{{\ell}_y}=1,
\end{align}
which leads to an update function for $\bm{\pi}$:
\begin{align} \label{eqn:update_leaf_cls}
\pi^{(t+1)}_{\ell_y}&=\frac{\sum_{i=1}^N\mathbf{1}(y_i=y)\rho({\ell}|\pi^{(t)}_{\ell_{y_i}},\mathbf{x}_i)}{\sum_{i=1}^N\rho({\ell}|\pi^{(t)}_{\ell_{y_i}},\mathbf{x}_i)}.
\end{align}
Eq.~\ref{eqn:update_leaf_cls} is the same as the update function for leaf nodes given
in~\cite{Ref:Kontschieder15}.

%



\ifCLASSOPTIONcompsoc
  \section*{Acknowledgments}
\else
  \section*{Acknowledgment}
\fi

This work was supported in part by the National Natural
Science Foundation of China No. 61672336 and in part by ONR N00014-15-1-2356. The authors
would like to thank Chenxi Liu, Siyuan Qiao and Zhishuai
Zhang for instructive discussions.

\ifCLASSOPTIONcaptionsoff
  \newpage
\fi



\bibliographystyle{IEEEtran}
\bibliography{mybib}
\end{document}